\documentclass{article}

\PassOptionsToPackage{numbers, compress}{natbib}

\usepackage[final]{neurips_2025}

\usepackage[utf8]{inputenc} 
\usepackage[T1]{fontenc}    
\usepackage{hyperref}       
\usepackage{url}            
\usepackage{booktabs}       
\usepackage{amsfonts}       
\usepackage{nicefrac}       
\usepackage{microtype}      
\usepackage{xcolor}         

\usepackage{amsmath}
\usepackage{amssymb}
\usepackage{braket}
\usepackage{algorithm}
\usepackage{algorithmic}
\usepackage{wrapfig}
\usepackage{booktabs}
\usepackage{multirow}
\usepackage{enumitem}

\usepackage{mathtools}
\usepackage{amsthm}

\usepackage{wrapfig}
\usepackage{braket}
\usepackage{caption}
\usepackage{graphicx}

\usepackage{siunitx} 
\usepackage{subcaption} 

\newtheoremstyle{boldremark}
  {\topsep}    
  {\topsep}    
  {\normalfont} 
  {}           
  {\bfseries}  
  {.}          
  { }          
  {}           

\theoremstyle{boldremark}

\title{Quantum Visual Fields with Neural Amplitude Encoding}

\author{
    \quad Shuteng Wang \quad\quad Christian Theobalt \quad\quad Vladislav Golyanik\thanks{Corresponding author's email: \href{mailto:golyanik@mpi-inf.mpg.de}{golyanik@mpi-inf.mpg.de}}\vspace{3pt}\\
    MPI for Informatics, SIC
}

\begin{document}

\maketitle

\begin{abstract}
    \noindent 
    Quantum Implicit Neural Representations (QINRs) have emerged as a promising paradigm that leverages parametrised quantum circuits to encode and process classical information.
    However, significant challenges remain in areas such as ansatz architecture design, the effective utility of quantum-mechanical properties, training efficiency, and the integration with classical modules.
    This paper advances the field by introducing a novel QINR architecture for 2D image and 3D geometric field learning, which we collectively refer to as Quantum Visual Field (QVF). 
    QVF encodes classical data into quantum statevectors using neural amplitude encoding grounded in a learnable energy manifold, ensuring meaningful Hilbert-space embeddings. 
    Our ansatz follows a fully entangled design of learnable parametrised quantum circuits, with quantum (unitary) operations performed in the real Hilbert space, resulting in numerically stable training with fast convergence. 
    QVF does not rely on classical post-processing---in contrast to the previous QINR learning approach---and directly employs measurements to extract learned signals encoded in the ansatz. 
    Experiments on a quantum hardware simulator demonstrate that QVF outperforms an existing quantum approach and competes with widely used classical foundational baselines in terms of visual representation accuracy across various metrics and model characteristics.
    We also show applications of QVF in 2D and 3D field completion and 3D shape interpolation, highlighting its practical potential. Project page: \url{https://4dqv.mpi-inf.mpg.de/QVF/}. 
\end{abstract} 

\section{Introduction} \label{sec:intro} 

\noindent 

Implicit neural representations~(INRs) have emerged as a powerful framework for continuously modelling signals via neural networks. 
They are widely used in image and 3D shape synthesis, as well as 3D reconstruction, among other fields of visual computing \cite{xie2022neural}. 
INRs map spatial (and also temporal) coordinates to corresponding signal values, enabling resolution-independent, memory-efficient, and differentiable representations; 
the signal encoding network $f_{\theta}$ with parameters $\theta$ is trained to minimise the reconstruction loss $\mathcal{L}(\theta)$ over sampled coordinates $\mathbf{x}$:   $\mathcal{L}(\theta) = \sum_{\mathbf{x} \in \mathcal{X}} \| f_{\theta}(\mathbf{x}) - \mathcal{S}(\mathbf{x}) \|^2 $, where \( \mathcal{X} \) denotes the sampled domain and $\mathcal{S}(\mathbf{x})$ is the signal value to be represented by $f_{\theta}$. 
As a remedy to the growing computational, memory and energy demand required by INR, recent work has explored the integration of quantum circuits into INR as a promising alternative to classical methods, with potential advantages in model compactness and learning efficiency~\cite{zhao2024quantum}. 
Quantum algorithms operate within Hilbert spaces, enabling superposition and entanglement of states that facilitate parallel processing beyond classical systems with comparable resource scales. 
Specifically, quantum machine learning (QML) models involving parameterised quantum circuits (PQCs) or \textit{ansatz} parameterise the evolution of quantum states through unitary transformations (implemented as quantum gate sequences), requiring a number of parameters that scale logarithmically with the Hilbert space dimensionality. 
Recent studies further reveal an intrinsic link between PQCs and Fourier-based learning mechanisms~\cite{schuld2021effect}, a critical feature for relieving biased learning or mitigating spectral bias common in classical neural networks for INRs~\cite{rahaman2019spectral}. 
Together, these insights suggest a pathway towards highly efficient and expressive QML models for visual computing. 
Despite their theoretical promise, quantum implicit neural representations (QINRs) remain heavily underexplored. 
The recently introduced QIREN approach \cite{zhao2024quantum} is, to our knowledge, among the first in the field that is closest to our work, designed for image representation, upsampling and generation. 
QIREN projects query coordinates into learnable Fourier features paired with a classical network decoder, explicitly linking it to Fourier neural architectures while, at the same time, overshadowing the quantum behaviour due to classical post-processing. 
\begin{wrapfigure}{r}{0.6\textwidth}
    \centering 
    \includegraphics[width = 0.6\textwidth]{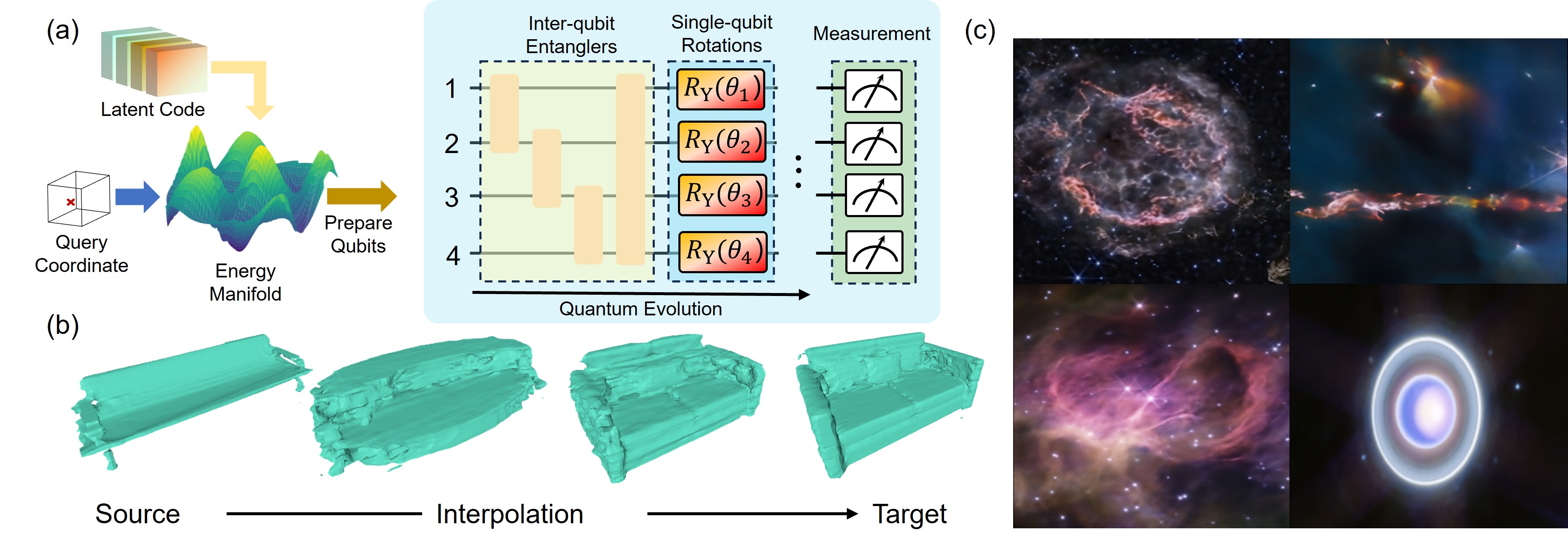}
    \caption{\textbf{Our learnable coordinate-based QVF model can represent various visual fields:} 
    (a) Schematic diagram of the architecture; 
    (b) Latent space interpolation of 3D signed distance fields \cite{shapenet2015}. 
    (c) 2D image representation of a moderate resolution~(400$\times$350 pixels) \cite{JWST};} 
    \label{Preview}
    \vspace{-15pt}
\end{wrapfigure}
In response to these limitations, we propose Quantum Visual Field (QVF), a novel coordinate-based QML model that leverages high-dimensional Hilbert spaces for lightweight and spectrally unbiased implicit visual field representations; see the scheme in Fig.~\ref{Preview}-(a). 
Rather than using heuristic classical-to-quantum data encoding methods ~\cite{weigold2020data,schalkers2024importance,bondarenko2020quantum,rathi20233d}---which (unreasonably) assume that handcrafted embeddings align with the inductive biases of quantum circuits---we introduce a novel learnable energy module that encodes non-linear data priors to align classical Euclidean and quantum Hilbert feature representations. 
Our carefully designed quantum circuit leverages quantum state evolution and entanglement between qubits 
to effectively explore the optimal pre-measurement quantum state representations. 
Specifically, the reachable Hilbert space is constrained in QVF for stable gradient flow and relief of issues such as barren plateaus, i.e., vanishing gradients arising due to Haar randomness\footnote{Haar randomness refers to the property of sampling quantum states uniformly at random from the Hilbert space according to the Haar measure. \cite{Haar1933}}, without compromising expressiveness.
The quantum circuit is measured to generate multi-dimensional signals, such as images or 3D geometries, or their collections (Fig.~\ref{Preview}-(b)) through conditioning on latent variables. 
In summary, the technical contributions of this paper include: 
\begin{itemize}
  \item[$\bullet$] QVF, a coordinate-based QML model for visual representation learning (2D images and 3D signed distance fields). 
  The QVF approach is designed for execution on quantum machine simulators or fault-tolerant gate-based quantum computers.  
  \item[$\bullet$] A non-linear neural scheme for encoding classical data into quantum statevectors. Our neural amplitude encoding is grounded in a learnable energy manifold ensuring meaningful Hilbert space embeddings. 
  \item[$\bullet$] An efficient PQC that processes entangled information within the real Hilbert subspace, explicitly designed for stable gradient feedback by bounding Haar randomness. 
\end{itemize} 

\noindent Unlike existing approaches \cite{zhao2024quantum}, QVF is a lightweight architecture with the exact structural configuration dynamically depending on the input data. 
We evaluate QVF and compare it to the main competitor, i.e.,~prior QINR method QIREN (on 2D image representation learning), and, additionally, several foundational classical INR baselines (for 2D image and 3D shape representation learning); experiments are performed on a high-end simulator of gate-based quantum hardware \cite{bergholm2018pennylane}. 
We show that QVF consistently competes and outperforms QIREN and other compared techniques. 
Moreover, QVF supports problem scales and applications beyond the reach of prior QINR frameworks, such as image inpainting, shape completion and latent space interpolation (Fig.~\ref{Preview}-(c)), taking a step towards unlocking quantum models in real scenarios. 

\raggedbottom
\begin{figure*}[t]  
    \centering
    \includegraphics[width = 1.0\textwidth]{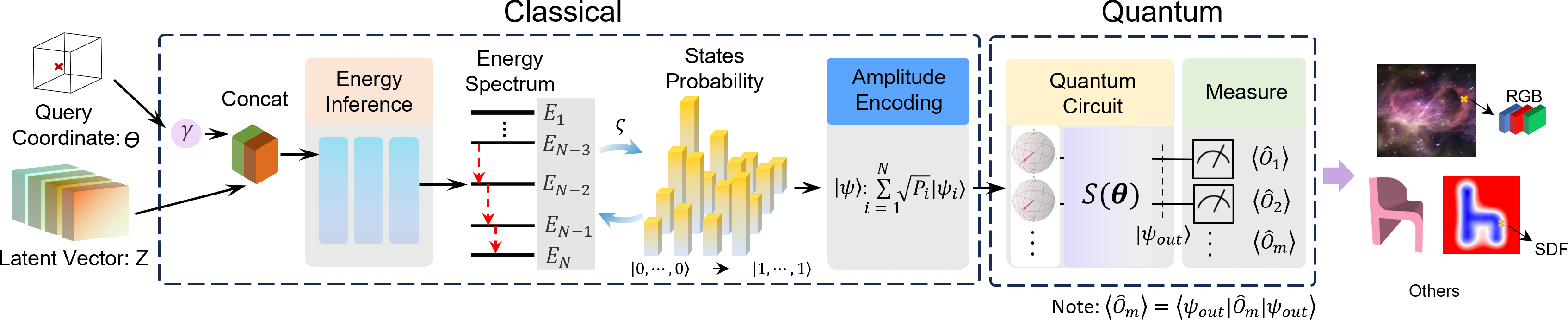} 
    \caption{\textbf{Overview of the proposed QVF model}, a QML framework for visual representation learning. 
    Query coordinates $\boldsymbol{\Theta}$ encoded using $\gamma$ (positional encoding) concatenated with the conditioning latent code $\boldsymbol{z}$ are used to infer the energy spectrum $\boldsymbol{E}$ of a quantum system, associated with Boltzmann-regulated statistical uncertainty $\boldsymbol{P}$. 
    The inferred statistical property is leveraged in encoding the classical data into quantum statevectors, which are subsequently processed by the parametrised quantum circuit $S(\boldsymbol{\theta})$. 
    Field properties are decoded probabilistically from projective circuit measurements.} 
    \label{quantum_framework} 
    \vspace{-15pt}
\end{figure*} 

\section{Related Work}

\newcommand{\cmark}{\textcolor{green}{\checkmark}}  
\newcommand{\xmark}{\textcolor{red}{\texttimes}}  

\noindent \textbf{Classical Neural 2D/3D Scene Representation.}
Neural networks serve as a basis for modern implicit scene representations, employing continuous function approximations to circumvent the constraints of discrete grid-based methodologies~\cite{molaei2023implicit,chen2017stylebank,tschernezki2022neural,chen2021learning,li20223d}. Initial breakthroughs utilised multi-layer perceptrons (MLPs) to establish coordinate-to-attribute mappings, as demonstrated in Chen et al.'s~\cite{chen2021learning} continuous implicit model for arbitrary-scale super-resolution.
The paradigm has since been extended to 3D representations, supplanting conventional voxel- and mesh-based approaches: DeepSDF~\cite{park2019deepsdf} achieves geometrically coherent surface reconstruction via learned signed distance fields (SDFs), while neural radiance fields (NeRF)~\cite{mildenhall2021nerf} introduce a volumetric scene representation parameterised by coordinate-based neural mappings of spatial coordinates and viewing directions to radiance and density, enabling photorealistic novel view synthesis.
Gate-based quantum computing offers great potential for fundamentally enhancing INRs. 

\noindent \textbf{Quantum-enhanced Computer Vision.} Growing interest in quantum computing for computer vision has established quantum-enhanced computer vision (QeCV) as an emerging research frontier \cite{meli2025quantum}. 
Current literature predominantly explores quantum annealers for combinatorial optimisation~\cite{golyanik2020quantum,birdal2021quantum,zaech2022adiabatic,farina2023quantum,benkner2021q,meli2025qucoop}, while tunable quantum circuits remain underexplored. 
Early works introduced foundational concepts such as quantum image denoising via localised convolutional operations~\cite{shiba2019convolution} and quantum convolutional neural networks (QCNNs) with mid-circuit measurements to emulate translational equivariance~\cite{cong2019quantum}.
Subsequent advances include hybrid quantum-classical architectures for 3D point cloud classification through voxelisation and quantum feature processing~\cite{baek20223d}, as well as quantum autoencoders for classical data compression via hand-crafted amplitude embeddings~\cite{rathi20233d}. 

Our work is inspired by 3D-QAE~\cite{rathi20233d}, who developed hand-crafted quantum amplitude embeddings \begin{wraptable}{r}{0.6\textwidth}
    \centering
    \vspace{-5pt}
    \scalebox{0.75}{%
    \begin{tabular}{@{} l|ccc @{}}
    \toprule
    \textbf{Characteristic} & \textbf{Ours} & \textbf{QIREN~\cite{zhao2024quantum}} & \textbf{3D-QAE~\cite{rathi20233d}} \\
    \midrule
    No heavy post-processing   &    \cmark & \xmark & \cmark \\
    Data encoding           & Neural AE   & Neural Angular      & AE (hand-crafted) \\
    Qubit budget           & logarithmic         & linear      & logarithmic \\
    Supported dimensions          & 2D/3D                   & 2D                      & 3D       \\
    Quantum hardware                  & Simulator               & Simulator               & Simulator \\
    \bottomrule
    \end{tabular}%
    }
    \caption{Comparative algorithmic analysis of related work. %
    ``AE'' denotes amplitude encoding. %
    }
    \vspace{-10pt}
    \label{tab:comparative}
\end{wraptable}
for encoding 3D point clouds.
However, their method, as acknowledged by their authors, suffers from limited scalability and underperforms classical models.
Another related work is QIREN by Zhao et al.~\cite{zhao2024quantum}, which employs the sandwich structure, i.e., a quantum circuit layer placed between classical pre- and post-processing.
They leverage the circuit's Fourier connections to project queries in the Fourier basis, followed by a classical dense layer for inference.
This draws parallels to classical positional or Fourier encodings, with the Fourier spectrum size growing exponentially.
While theoretically motivated, its practical utility is debated as heavy classical postprocessing reduces the quantum component to a feature generator.
In contrast, our framework avoids such post-processing and
more heavily relies on the ansatz; 
see an algorithmic comparison in Table~\ref{tab:comparative}. 
We use a learnable, energy-based Boltzmann-regulated amplitude encoding, which is a critical step towards unlocking the potential of quantum computing as demonstrated empirically in Sec.~\ref{sec:Experiments}. 
At the same time, our carefully designed ansatz evolves the encoded data and ensures robust gradient feedback. 

\section{Review: QML, its Unitary Nature and Fourier Structure} \label{s:Fourier_structure}

This paper assumes familiarity with quantum computing and its notations. 
For convenience, we provide a refresher in 
App.~\ref{s:background}. 
QML leverages parametrised unitary quantum operations on encoded data $|\psi(x)\rangle = \sum_j \psi_j(x) |j\rangle$ to learn functions typically expressed as expectation values $f(x) = \langle \psi(x) | \hat{O} | \psi(x) \rangle$, where $\hat{O}$ is a Hermitian observable ($\hat{O} = \hat{O}^\dagger$).
The unitary nature of quantum evolution ($U^\dagger U = U U^\dagger = I$) preserves inner products and norms, ensuring that the spectral components of the encoded data are transformed by quantum circuits.
The spectral decomposition of $\hat{O} = \sum_k \lambda_k |e_k\rangle\langle e_k|$ reveals a Fourier-like structure in $f(x) = \sum_k \lambda_k |\langle e_k | \psi(x) \rangle|^2$, where the projections $\langle e_k | \psi(x) \rangle$ act as Fourier coefficients and the eigenvalues $\lambda_k$ relate to accessible frequencies. 
This fundamental property is a consequence of quantum mechanics: unitary transformations preserve spectral components, ensuring that even complex quantum circuits inherently operate in a frequency domain. As a result, the expressivity of QML models is directly linked to their accessible frequency components, influencing their ability to generalise and learn structured data representations. 

\section{Our QVF Approach}\label{sec:QVFs} 

This section introduces the proposed QVF model, i.e., a QINR for learning visual representations and their collections; 
see Fig.~\ref{quantum_framework} for its architecture. 
QVF takes query coordinates $\boldsymbol{\Theta}$ and an optional latent variable $\boldsymbol{z}$ (in the case more than one visual field needs to be represented) as inputs and produces 2D or 3D field properties $s$. 
We introduce encoding classical data into quantum states using neural amplitude encoding in Sec.~\ref{ss:energy}, while the quantum circuit design and measurement are detailed in Sec.~\ref{s:circuit}. 
Sec.~\ref{s:training} provides training details and applications supported by QVF. 

\setlength{\columnsep}{10pt} 

\subsection{Amplitude Encoding with Neural Embeddings} \label{ss:energy} 

\noindent Our parametrised energy-based embedding of classical data $x$ into quantum states $\ket{\psi_{in}(x)}$ generalises widely used hand-crafted amplitude encoding (AE)~\cite{rathi20233d,schleich2024quantum}. 
AE enables an exponentially compact encoding of $N = 2^n$ classical values into probability amplitudes of $n$ qubits by leveraging quantum superposition. 
Notably, this implies that AE induces \textit{exponentially-many} random Fourier features due to the inherent periodicity of quantum state phases. 
The fundamental limitation of hand-crafted AE stems from its \textit{a priori, possibly biased} prepared quantum states, which poses a risk of misalignment with subsequent quantum evolution or suboptimal utilisation of task-specific data Fourier priors. 
Hence, we propose a data-driven approach for AE that learns the optimal quantum state density \( \hat{\rho}_{\text{opt}}(x) \), directly from data, i.e., for QINR in our case. 
We restrict the process on pure quantum states satisfying \( \text{Tr}(\hat{\rho}(x)^2) = 1 \). 
Drawing upon the fundamental energy-probability duality inherent in physical systems (e.g.,~in statistical and quantum mechanics), we infer the conditional energy spectrum $\boldsymbol{E}$ of a given visual representation and transform it into a probability distribution $\boldsymbol{P}$ subsequently encoded as qubit state amplitudes \(\alpha_i \in \mathbb{C}\) residing in the complex Hilbert space \(\mathcal{H}\). 
Our encoding introduces \textit{non-linearity into the quantum evolution} while reserving the full repertoire of quantum processing and measurements\footnote{transformations in quantum circuits before measurement are linear, which is often seen as a key limitation in ansatz expressivity}. 
For energy inference, we employ a minimal dense MLP $f(x=\{\boldsymbol{\Theta}, \textbf{z}\}):\gamma(\boldsymbol{\Theta}) \times \textbf{z} \rightarrow \boldsymbol{E}$ activated by ReLU; we use positional encoding to accelerate learning \cite{rahaman2019spectral,mildenhall2021nerf}:
\begin{equation} \label{eq:2}
\gamma(\boldsymbol{\Theta}) = ( 
   \cdots , \sin(2^{L-1}\pi \boldsymbol{\Theta}),\cos(2^{L-1}\pi \boldsymbol{\Theta}), \cdots). 
\end{equation}
$\boldsymbol{\Theta}$ denotes the field query coordinate while $\boldsymbol
{z}$ represents the latent code in the case of learning visual field collections.
The inferred $\boldsymbol{E} = f(\gamma(\boldsymbol{\Theta}), \textbf{z})$ 
is leveraged to derive the Boltzmann-regulated
$ \boldsymbol{P} $ of the quantum system; the Gibbs-Boltzmann framework $\varsigma$ serves as an inductive embedding bias for encapsulating thermodynamic uncertainty, enabling the realisation of Gibbs quantum states~\cite{amin2018quantum, ball2023boltzmann}. 
We next formulate \(\boldsymbol{P} = [P_i] \), $i \in \{1, \hdots, N\}$ through the construction of a discretised energy landscape \(\boldsymbol{E}\), derived from the Gibbs canonical ensemble: 
\begin{equation}
\label{eq:3}
\boldsymbol{P} = \frac{\exp\left(-\beta \boldsymbol{E}(\boldsymbol{\Theta}, \textbf{z})\right)}{\mathcal{Z}},\,\text{where} 
\end{equation}
\begin{equation} \label{eq:partition_function}
\mathcal{Z} = \int \exp\left(-\beta \boldsymbol{E}(\boldsymbol{\Theta}, \textbf{z}\right) d\boldsymbol{\Theta} \approx \sum_{j=1}^{N} \exp\left(-\beta E_j(\boldsymbol{\Theta}, \textbf{z})\right), 
\end{equation}
with \(\beta = (k_B T)^{-1}\) representing the inverse temperature. 
The quantum state amplitudes \(\alpha_i \in \mathbb{C}\) residing in the complex Hilbert space \(\mathcal{H}\) are characterised by their complex phases \(\phi_i = \arg(\alpha_i)\) which are arbitrary within the interval \([0, 2\pi)\); it is subjected to the normalisation condition \(\lVert \alpha_i \rVert_2^2 = P_i\) with \(\sum_{i=1}^N P_i = 1\). 
Finally, the quantum states $\ket{\psi_{in}(\boldsymbol{\Theta}, \textbf{z})}$ 
encoding the query coordinates for our input fields 
are prepared as follows: 
\begin{equation} \label{eq:4}
    \begin{aligned}
        &\ket{\psi_{in}(\boldsymbol{\Theta}, \textbf{z})} = \sum_{i=1}^{N} \alpha_i \ket{\psi_i}, \quad \alpha_i = \sqrt{P_i} e^{i \cdot \operatorname{arg}(\alpha_i)}, \\
        \quad &\hat{\rho}(\boldsymbol{\Theta}, \textbf{z}) = \ket{\psi_{in}(\boldsymbol{\Theta}, \textbf{z})}\bra{\psi_{in}(\boldsymbol{\Theta}, \textbf{z})} = \sum_{i,j = 1}^{N} \alpha_i \alpha_j^+ \ket{\psi_i}\bra{\psi_j}.
   \end{aligned}
\end{equation}
$\ket{\psi_{i}}$ is the computational basis,  
$\hat{\rho}(\boldsymbol{\Theta}, \textbf{z})$ is the density distribution of $\ket{\psi_{in}(\boldsymbol{\Theta}, \textbf{z})}$ and ``$(\cdot)^+$'' denotes the adjoint. 
We then theoretically analyse data encoding effects on the model expressiveness.

\noindent {\textbf{Lemma 1} \textit{Energy inference exhibits functional equivalence to determining optimal non-linear input-dependent frequency spectrum embedded within variational quantum circuits, defining the model's inherent expressiveness.}}

\noindent As demonstrated by Schuld et al.~\cite{schuld2021effect}, variational circuits of the form $U(x) = W^2 g(x) W^1$ admit a truncated Fourier-type expansion when measuring circuit expectation values $\braket{\hat{M}}$: 
\begin{equation} \label{energy_asso_freq_spectrum}
\braket{\hat{M}} = \bra{0} {W^1}^{\dagger}{g(x)}^{\dagger} \tilde{M} g(x) W^1 \ket{0} =  \sum_{w \in \Omega} c_w e^{iwx}, 
\end{equation}
where $W^1$ and $W^2$ are arbitrary unitary matrices. 
The effective measurement operator is defined as: $\tilde{M} = {W^2}^{\dagger} \hat{M} {W^2}$, while $g(x)$ serves as data encoding modules applied to the physical system.
Notably, unlike Schuld et al.~\cite{schuld2021effect}, where encoding gate analysis is restricted to Pauli gates, $g(x)$ encompasses more general quantum operations.
Prepared input quantum state $\ket{\psi_{in}(x)}$ can be equivalently expressed as  
\begin{equation}  
\ket{\psi_{in}(x)} = g(x) W^1 \ket{0},  
\end{equation}  
establishing a direct correspondence between inferred energy landscape and the multi-dimensional frequency spectrum $\Omega$, with dependencies encoded in the learnable energy inference framework.
The optimal effective measurement basis, given by \(\hat{M}_{\text{opt}} = {W^2_{\text{opt}}}^{\dagger} \hat{M} {W^2_{\text{opt}}}\), along with the heuristic learnable circuit design \(\hat{S}(\boldsymbol{\theta})\), which approximates \(W^2_{\text{opt}}\), will be introduced in the next sections.

\setlength{\columnsep}{10pt}
\subsection{Our Parametrised Quantum Circuit} \label{s:circuit} 

\noindent \begin{wrapfigure}{r}{0.2\textwidth}
    \vspace{-15pt}
    \includegraphics[width = 0.2\textwidth]{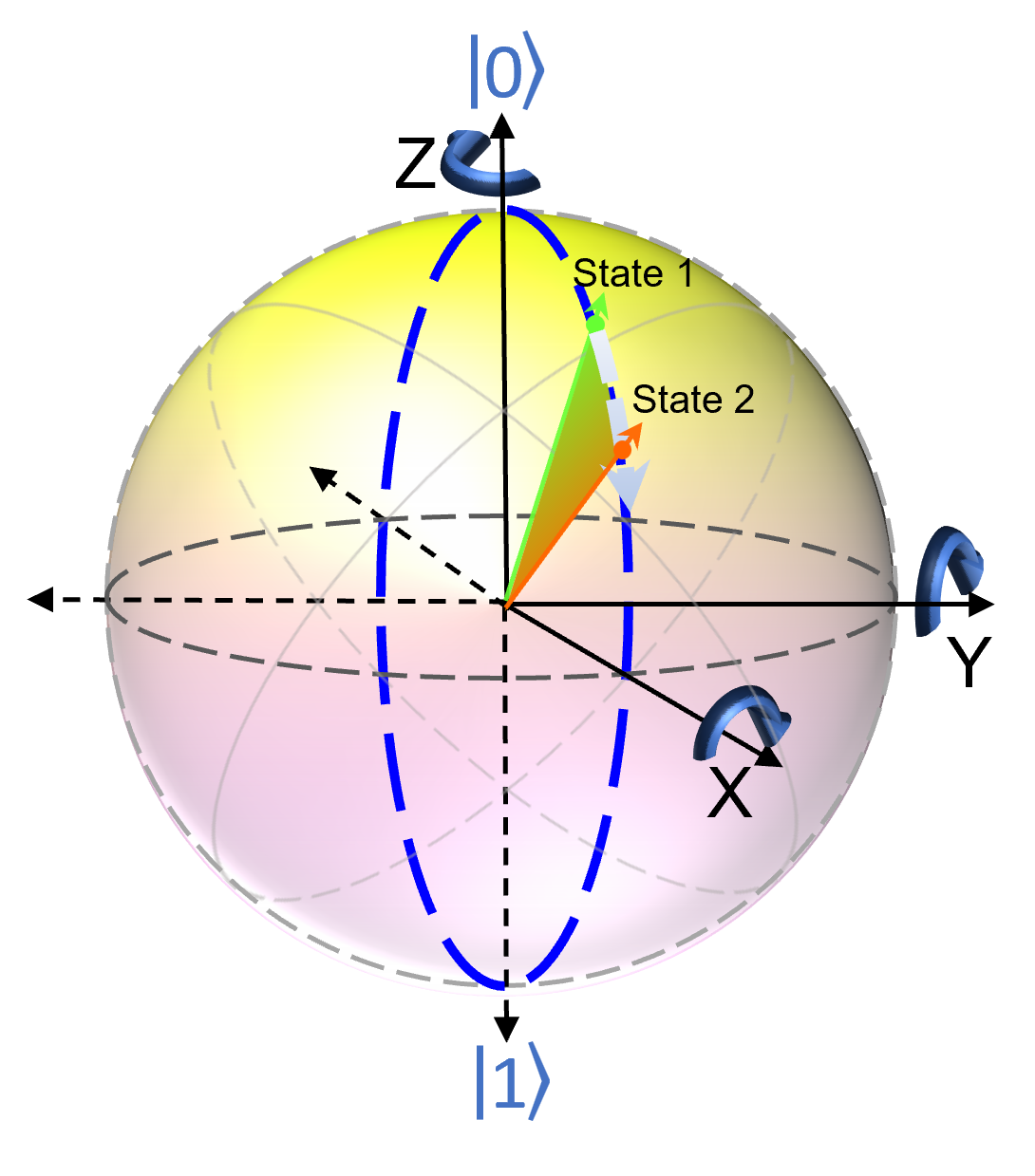} 
    \caption{Representative pure qubit state transitions on the Bloch sphere.}
    \vspace{-15pt}
    \label{qubit state visualization}
\end{wrapfigure}Once the classical data is encoded into $\ket{\psi_{in}(\boldsymbol{\Theta}, \textbf{z})}$, it is processed by our learnable PQC or ansatz $\hat{S}(\boldsymbol{\theta})$ within a high-dimensional Hilbert space. 
Our goal is a compact and expressive PQC for QINR learning. 
As unrestricted traversal of the Hilbert space can induce training instabilities, we therefore constrain the ansatz $\hat{S}(\boldsymbol{\theta})$ to the manifold of real-valued unitaries, constructed from Pauli-Y rotations and entangling gates to ensure efficient training of QVF.
This design choice avoids the imaginary components introduced by Pauli-X and -Z rotations, which would otherwise permit unconstrained exploration of the full, complex Hilbert space and hinder training.
Our PQC architecture is analogous to classical densely-connected neural networks in the sense that it contains alternating layers of parametrised single-qubit Pauli-Y rotations and entangling operations, supporting highly correlated, non-local quantum states that cannot be decomposed into a tensor product of individual qubits, which enables parallel information processing.
This design provides critical benefits for QINR learning: (1) it confines state evolution to the manifold of real-valued unitaries, eliminating redundant parameter dimensions that 
facilitate scrambled quantum states and barren plateaus; (2) it naturally discards complex phase information while \textit{preserving all measurement-relevant quantities, i.e., universality,} for certain basis observations such as the Pauli-$Z$ observable, as the complex phase factors cancel out and become irrelevant when computing $Z$-basis probabilities; and (3) it maintains full expressivity while significantly simplifying the optimisation landscape. 
Additionally, our design necessitates the enforcement of a zero complex phase in our encoded quantum data, formally expressed as \(\operatorname{arg}(\alpha_i) = 0\), with schematic Bloch sphere dynamics depicted in Fig.~\ref{qubit state visualization}.
Once $\ket{\psi_{in}(\boldsymbol{\Theta}, \textbf{z})}$ has been transformed by $\hat{S}(\boldsymbol{\theta})$, we extract the visual field attributes 
encoded in our QINR using projective measurements of the final quantum states. 

\noindent \textbf{Multi-dimensional Measurement}. To extract an \( m \)-dimensional representation (\(m{\leq}n \)) from $\hat{S}(\boldsymbol{\theta})$, we implement local Pauli projective measurements on the first \( m \) qubits, effectively tracing out the remaining \( n{-}m \) qubits. 
The corresponding family of local measurement operators \(\{\hat{O}_i\}\) is formally defined over the \( n \)-qubit Hilbert space as:
\begin{equation}
    \hat{O}_i = \left( \otimes_{k=1}^{i-1} \mathbb{I}_k \right) \otimes \sigma^Z_i \otimes \left( \otimes_{l=i+1}^{n} \mathbb{I}_l \right), \quad i \in \{1,\ldots,m\},
\end{equation}
where \(\mathbb{I}_k\) denotes the identity operator acting on the \( k \)-th qubit, preserving its quantum state within the tensor product. 
The operator \(\sigma^Z_i = \ket{0}\!\bra{0}_i - \ket{1}\!\bra{1}_i\) represents the Pauli-Z observable applied to the \( i \)-th qubit. 
Local measurements help guarantee robust gradient feedback \cite{cerezo2021cost,thanasilp2023subtleties} for the circuit. 
The output of $\hat{S}(\boldsymbol{\theta})$ is defined as the expectation value of finite-shot circuit measurements (App.~\ref{s:image_represent_finite_samples} analyses the influence of circuit measurements on the extracted image quality). 
This expectation value can be expressed as $V_{\operatorname{inf}}$, which is defined in the asymptotic limit as the number of shots approaches infinity: 
\begin{equation} \label{eq:6}
V_{\operatorname{inf}}(\boldsymbol{\Theta}) = \operatorname{Tr}(\hat{\rho} (\boldsymbol{\Theta}) \hat{M}(\boldsymbol{\theta})), \; \hat{M}(\boldsymbol{\theta)} = \hat{S}(\boldsymbol{  \theta  })^{\dag} \hat{O} \hat{S}(\boldsymbol{  \theta  }). 
\end{equation}
$\hat{M}(\boldsymbol{\theta})$ represents a parametrised measurement basis employed to approximate the optimal measurement basis $\hat{M}_{\text{opt}}$ via unitary quantum evolutions of 
our fully-entangled circuit $\hat{S}(\boldsymbol{  \theta })$. 
Similar to classical universal approximation theory, quantum circuits with sufficient depth can approximate arbitrary unitary transformations.
The Solovay--Kitaev Theorem provides a rigorous theoretical upper bound on the number of quantum gates required to approximate an arbitrary unitary operation to a given precision $\epsilon$, given by \(O(4^n \log^4 (1/\epsilon))\).
Circuit depth $\mathcal{J}$---analogous to the number of layers in classical neural networks---and the corresponding total number of constituent unitary quantum gates are hyperparameters of our $\hat{S}(\boldsymbol{  \theta })$. 
The local measurement $V(\boldsymbol{\Theta})_i$ corresponding to the $i$-th qubit is injectively mapped 
to the corresponding dimension of the target field, requiring the qubit number $n{\ge}m$. 
We next detail the end-to-end training protocol of our model under the Bayesian framework. 

\subsection{Training Details and Applications} \label{s:training}

\begin{wrapfigure}{r}{0.5\linewidth}
    \vspace{-15pt}
    \includegraphics[width=1\linewidth]{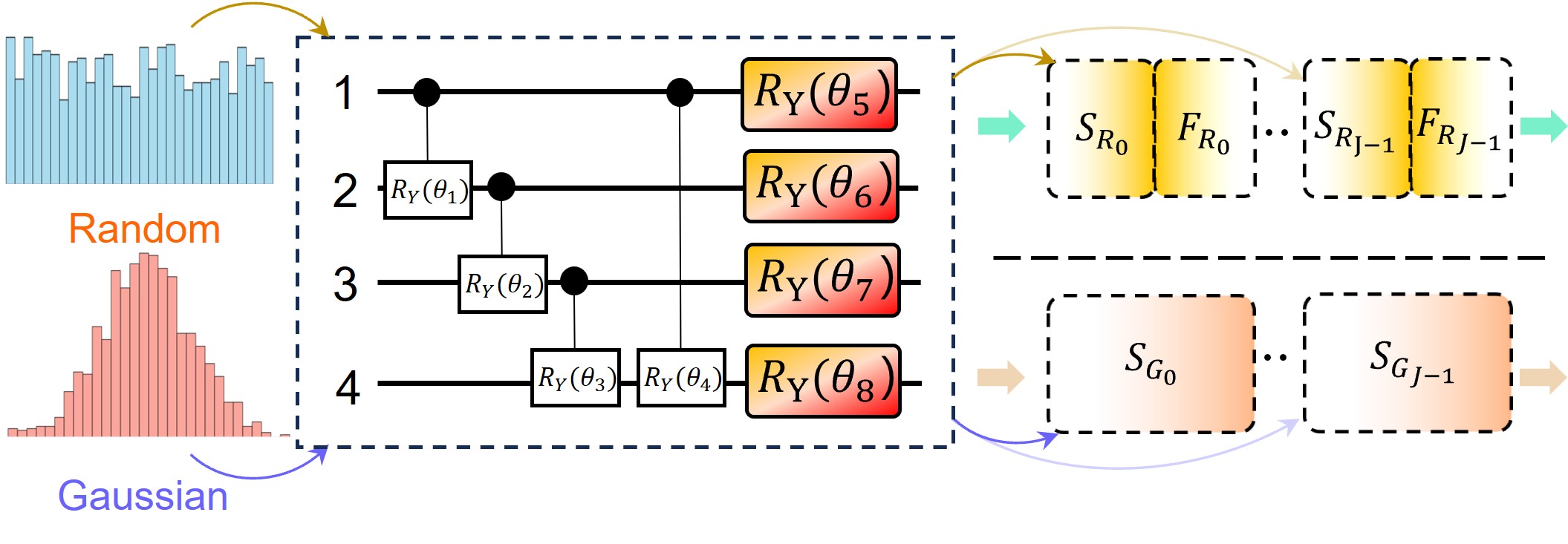}
    \caption{\textbf{QVF Initialisation}: Schematic circuit module initialised with identity (top) and  Gaussian (bottom) schemes.} 
    \label{quantum_circuit}
    \vspace{-15pt}
\end{wrapfigure}

\noindent \textbf{Initialisation of $\hat{S}(\boldsymbol{\theta})$.} We incorporate established PQC initialisation strategies, i.e.,~identity~\cite{grant2019initialization} and Gaussian~\cite{zhang2022escaping}; see Fig.~\ref{quantum_circuit} for the architectural implications. 
For identity initialisation,
 each circuit layer $\hat{S}(\boldsymbol{\theta})_j$ at depth $j \in \{1, \cdots, J\}$ can be expressed as a sub-circuit $S_R$ followed by its Hermitian adjoint $F_R$. 
Then, $\hat{S}(\boldsymbol{\theta})_j$ is constructed by assigning \(\theta_k \sim \mathcal{U}[0, 2\pi)\) to $S_R$, which also initialised as $F_R$ per definition.
This ensures that the composite operation $\hat{S}(\boldsymbol{\theta})_j = S_R F_R$ is equivalent to a zero circuit depth (identity circuit) before training. 
Note that while the initial configuration enforces $S_R F_R = I$, this constraint is not maintained during optimisation \cite{grant2019initialization}.
For the Gaussian initialisation, 
trainable parameters for $\hat{S}(\boldsymbol{\theta})_j = S_{G_j}$ are, instead, sampled from a zero-mean Gaussian distribution with the variance coupled to the circuit depth, i.e.,~\(\theta_k \sim \mathcal{N}(0, \sigma^2(\mathcal{J}))\)~\cite{zhang2022escaping}.
The overall circuit architecture \( \hat{S}(\boldsymbol{\theta}) \) is obtained by concatenating \( \mathcal{J} \) blocks $\hat{S}(\boldsymbol{\theta})_j$ 
such that the overall unitary transformation is given by \(\hat{S}(\boldsymbol{\theta}) = \prod_{j=1}^{\mathcal{J}} \hat{S}(\boldsymbol{\theta})_j\).
Note that architectural homogeneity across blocks is maintained, preserving systematic exploration of the unitary space \(\mathcal{U}(2^n)\).

\noindent\textbf{QVF Training.} Consider dataset $X$ composed of $W$ distinct visual fields denoted by $X_i$, for $i \in \{1, ..., W\}$. Each data field $X_i$ encapsulates physical field properties $s_i^j$, such as pixel values in images or SDF for geometric representations, sampled at specific spatial coordinates $\Theta_i^j$; here, index $j$ denotes the sample index per field. 
The relationship between spatial coordinates and physical properties is defined by a function $f$, such that sampled points within each field are given by:
\begin{equation}\label{eq:9}
    X_i = \{(\Theta_i^j, s_i^j) | s_i^j = f(\Theta_i^j), j \in \{0, 1, ..., M\}\}, 
\end{equation}
where \( M \) is the number of samples per field. 
Crucially, each data field $X_i$ is associated with a unique latent code $\boldsymbol{z_i}$. 
The training objective is to maximise conditional probability distribution $p_{\boldsymbol{\theta}}(\mathbf{s} | \boldsymbol{\Theta})$; $\boldsymbol{\theta}$ represents trainable parameters of a QVF: 
\begin{equation}
    p_{\boldsymbol{\theta}}(\mathbf{s} | \boldsymbol{\Theta}) = \sum_{i,j} p_{\boldsymbol{\theta}}(s_i^j | \Theta_i^j, \boldsymbol{z_i}) p(\boldsymbol{z_i}).
\end{equation}
Under a sufficiently large number of i.i.d.~quantum circuit measurement shots, the conditional likelihood 
$p_{\boldsymbol{\theta}}(s_i^j | \Theta_i^j, \boldsymbol{z_i})$ can be approximated by a Gaussian distribution. 
This statement is supported by the Central Limit Theorem (CLT), which establishes the asymptotic normality of the sum (or average) of numerous i.i.d.~random variables with finite variance. 
Consequently, $p_{\boldsymbol{\theta}}(s_i^j | \Theta_i^j, \boldsymbol{z_i})$ can be approximated as 
\begin{equation}
    p_{\boldsymbol{\theta}}(s_i^j | \Theta_i^j, \boldsymbol{z_i}) \approx \exp \left(-\mathcal{L}^2 \big(V (\boldsymbol{z_i}, \Theta_i^j; {\boldsymbol{\theta}}), s_i^j \big) \right),
\end{equation}
where $\mathcal{L}(\cdot)$ represents the loss function that quantifies the discrepancy between the output of the circuit $V (\boldsymbol{z_i}, \Theta_i^j; {\boldsymbol{\theta}})$ and the observed physical property $s_i^j$.
Model training can, therefore, be formulated as maximising this conditional likelihood under a Bayesian framework. 
To ensure a smooth representation transition in the latent space, the prior distribution over \( \boldsymbol{z_i} \) is softly penalised to follow a smooth distribution; an isotropic zero-mean multivariate Gaussian distribution is a reasonable choice as adopted by Park et al.~\cite{park2019deepsdf}.
The loss function \( L_{\boldsymbol{\theta}, \boldsymbol{z}} \), minimised via training over all learned \( W \)  fields $\{s_i | i = {1, \hdots, W} \}$ with \( M \) samples per $s_i$, is formulated as: 
\begin{equation} \label{eq:11} 
\scalebox{0.92}{$
L_{\boldsymbol{\theta}, \boldsymbol{z}}(\boldsymbol{\Theta}, \boldsymbol{s}) = \sum_{i, j=1}^{W,M} (\mathcal{L} (V (\boldsymbol{z_i},\Theta_i^j; {\boldsymbol{\theta}}), s_i^j) + \gamma \|\boldsymbol{z_i}\|_2).
$}
\end{equation} 
QVF undergoes end-to-end training: classical parameters are updated via gradient descent, while quantum parameters are optimised using the parameter-shift rule \cite{mitarai2018quantum}. 

\noindent \textbf{Usage and Applications.} 
Once trained, we can query QVF for the encoded 2D or 3D representations in a coordinate-based manner. 
We can also infer with partial samples, enabling applications such as image inpainting and partial shape completion through latent space optimisation. 
Using Maximum-a-Posteriori (MAP) estimation, we identify a latent code \( \hat{\boldsymbol{z}} \) that maximises agreement with the input partial observation \( \hat{X}_i \) while keeping the pre-trained model fixed: 
\begin{equation} \label{eq:12}
    \hat{\boldsymbol{z}} = \underset{\boldsymbol{z}}{\text{argmin}}{\sum_{(\Theta_j,s_j)} \mathcal{L} (V (\boldsymbol{z_i},\Theta_j; {\boldsymbol{\theta}}), s_j) + \gamma ^2 ||\boldsymbol{z}||_2}. 
\end{equation} 
\noindent\textbf{Algorithmic Summary.} 
We summarise the QVF training protocol in Algorithm~\ref{alg:qvf_protocol} in the Appendix. 

\section{Experimental Evaluation} \label{sec:Experiments} 

\noindent We experimentally evaluate our QVF 
for learning visual field representations, encompassing both 2D images and 3D geometries, while systematically analysing its generalisation in the sense of signal interpolation and the ability to handle missing and occluded regions. 
We use 1) images from the CIFAR-10~dataset~\cite{krizhevsky2009learning} and high-resolution images with rich spectral details~\cite{JWST}, and 2) 3D shapes from the ShapeNet~\cite{shapenet2015} dataset. 
We report widely used metrics averaged over three repetitions.

\noindent \textbf{Implementation Details.}
We empirically evaluate the model on a noiseless high-end simulator: \textit{default.qubit.torch}, provided by PennyLane~\cite{bergholm2018pennylane} with an A100 GPU. 
We employ Adam optimisation \cite{kingma2014adam} with an initial learning rate of $\eta = 10^{-3}$, subject to a learning rate scheduler that triggers upon plateauing with a window size of $50$ epochs (scaling $\eta$ by $0.9$).  
The number of epochs is set to $5k$, and $\gamma = 10^{-3}$ in Eq.~\eqref{eq:11}. 

\noindent \textbf{Hardware and Efficiency.} 
Absence of large-scale, fault-tolerant quantum hardware forces contemporary QML models to rely on exponentially expensive simulators run on classical hardware; see Table~\ref{tab:comparative}.
For a circuit with $n$ qubits of depth $\mathcal{J}$, the computational complexity on a classical noiseless simulator, without acceleration, is given by $O(2^{c\cdot n} \mathcal{J})$ where $c$ is a constant that depends on the specific simulation method employed.

\newpage
\subsection{Circuit Trainability} 
\label{ssec:Trainability} 

\begin{wrapfigure}{r}{0.5\linewidth}
    \vspace{-12pt}
    \includegraphics[width=1\linewidth]{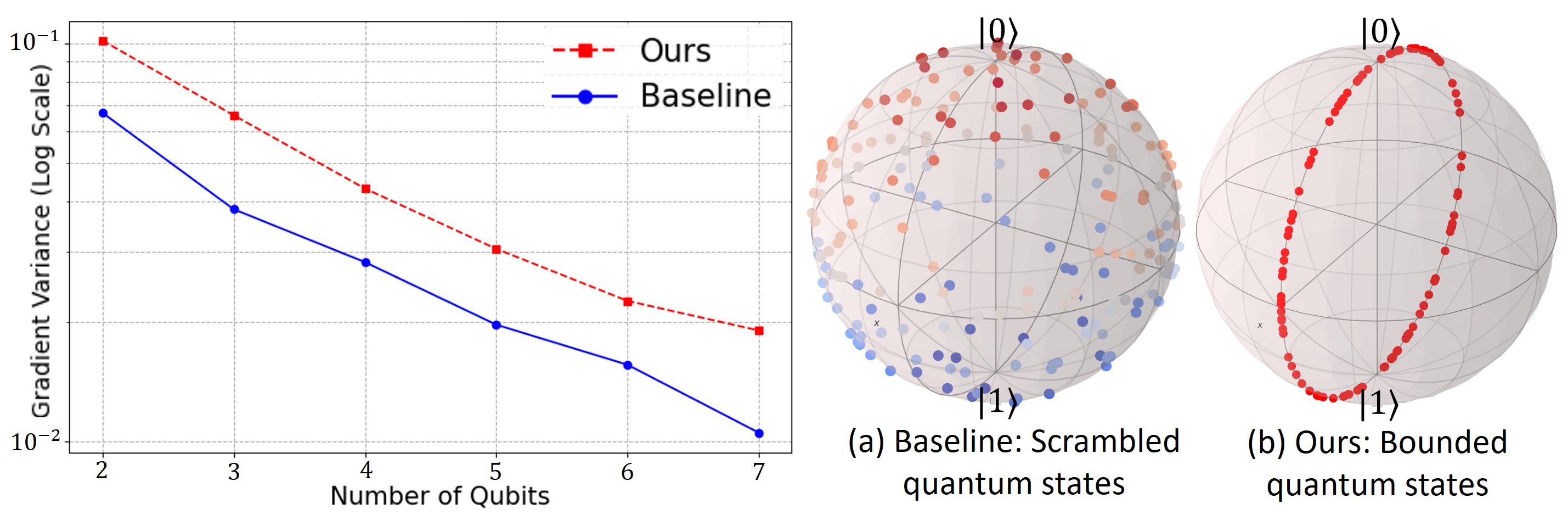}
    \caption{(Left:) comparison of the gradient variance ($y$-axis has a log scale); (right:) visualisation of reachable quantum states.}
    \label{gradient_behaviour}
    \vspace{-10pt}
\end{wrapfigure} We experimentally show that constraining quantum transformations in $\hat{S}(\boldsymbol{\theta})$ to real-valued unitary operations (resulting in bounded Haar randomness) helps with gradient flow. 
As quantum circuit parameters are inherently 
periodic within $[0, 2\pi)$, we evaluate the gradient flow by uniformly sampling parameters within this range and quantifying its expectation value. 
Due to the zero-mean nature of the expected loss gradient (see App.~\ref{barren_pleatau}), the vanishing gradient phenomenon is governed by the variance decay rate. 
We, therefore, quantify its variance $\operatorname{Var}_{\text{grad}}$ as 
\begin{equation}
    \operatorname{Var}_{\text{grad}} = E_{\theta \sim U([0,2\pi))} \left[\operatorname{Var}\left(\left\{\frac{\partial}{\partial \theta_k} \langle \hat{M} \rangle_{\theta}\right\}_{t=1}^T\right)\right], 
\end{equation}
where ``$\operatorname{Var}(\cdot)$'' denotes the variance operator 
and $T = 500$ is the number of samples to evaluate the expectation. 
$\langle \hat{M} \rangle_{\theta}$ is the expectation value of $\hat{S}(\boldsymbol{\theta})$, and  
$k$ iterates over ansatz parameters. 
Fig.~\ref{gradient_behaviour} reports $\operatorname{Var}_{\text{grad}}$ for the increasing number of qubits for two ansatze, i.e., of our QVF and QIREN~\cite{zhao2024quantum} with a strongly-entangled quantum circuit which allows scrambled (i.e.,~non-restricted) quantum states in the Hilbert space. 
We observe that our ansatz with bounded Haar randomness maintains a stronger gradient flow, which is crucial for its trainability and efficient representation learning. 
\subsection{2D (Image) Representation Learning} \label{ssec:2D_comparisons}

\begin{wrapfigure}{r}{0.6\textwidth}
    \vspace{-10pt}
    \includegraphics[width=\linewidth]{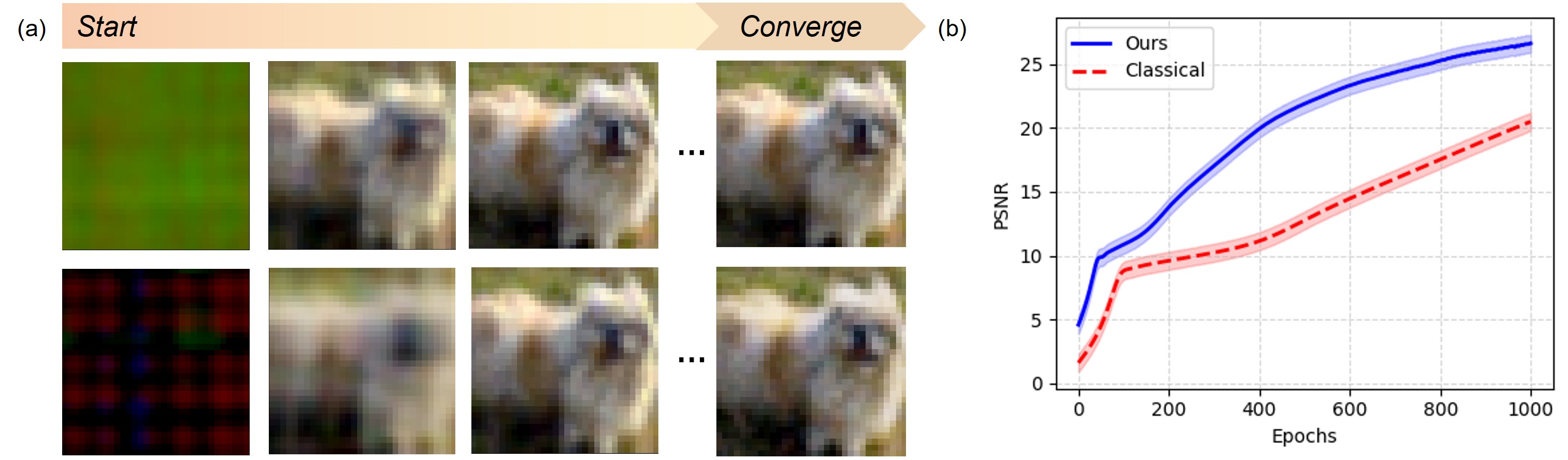}
    \captionof{figure}{(a) Reconstructed images during training: (top) our QVF, (bottom) classical model; (b) PSNR curves.
    }
    \label{dog_pic}
    \vspace{-13pt} %
\end{wrapfigure} We evaluate image representation learning with QVF and start with single images. 
We first compare QVF to a classical model that takes the architecture consistent with QVF's classical energy inference module; QVF has an overhead of $170$ parameters due to its ansatz. 
This implies that the differences in the learning behaviour and the final representation accuracy are \textit{predominantly} due to the inductive bias of the quantum ansatz, isolating influences from external factors.
Fig.~\ref{dog_pic}-(a) visualises reconstructed images during training of QVF; Fig.~\ref{dog_pic}-(b) plots the learning curves (PSNR) for the first thousand training epochs and, thus, highlights the differences in the training progression. 
Similar observations are made for other trials during the evaluation; QVF significantly accelerates learning high-frequency signals while performing on par with the classical method in the low-frequency regions. 

\begin{wrapfigure}{r}{0.4\textwidth}
    \vspace{-25pt} 
    \begin{minipage}[t]{0.4\textwidth}
        \includegraphics[width=\linewidth]{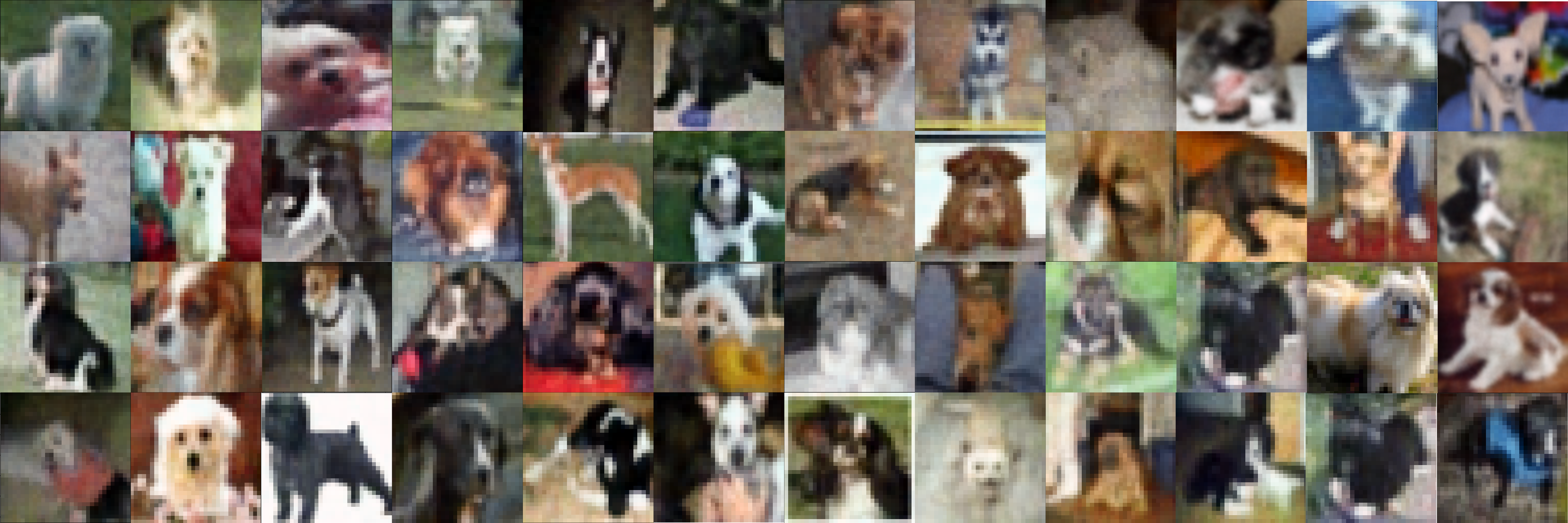}
        \captionof{figure}{Visualisation of the reconstructed images.}
        \label{more_dogs}
    \end{minipage}
    \hfill
    \begin{minipage}[t]{0.4\textwidth}
        \includegraphics[width=\linewidth]{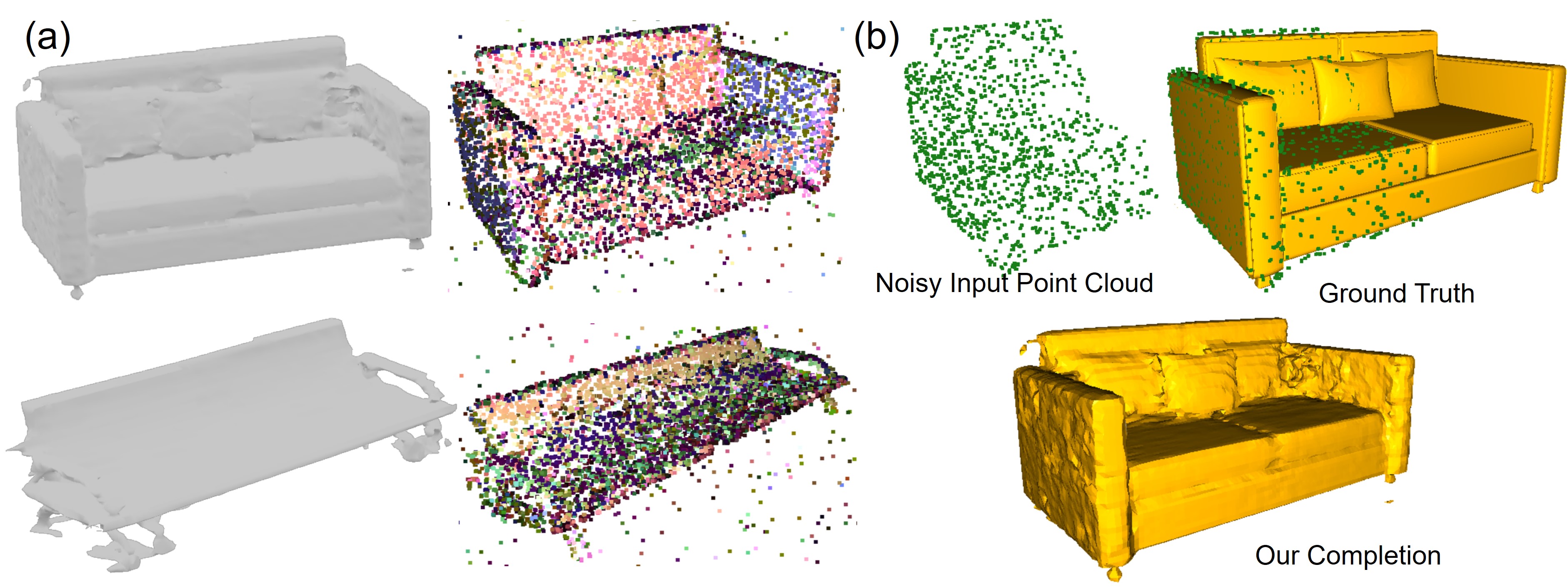}
        \captionof{figure}{(a): Geometry representation using QVF with latent-space-conditioned SDF inference; (b): Shape completion from partial inputs.}
        \label{chairs}
        \vspace{-5pt} 
    \end{minipage}
    \vspace{-15pt} 
\end{wrapfigure} We also experiment with the hand-crafted encoding strategy of Rathi et al.~\cite{rathi20233d}, which does not result in a recognisable representation upon convergence---an observation consistent with their results. 
This validates our design and, especially, the necessity of a learnable energy embedding. 
We then benchmark QVF against QIREN~\cite{zhao2024quantum}, the most closely related QINR approach. 
While QIREN employs a quantum ansatz sandwiched between classical layers, QVF uses a classical component for data encoding only. 
We evaluate QVF and QIREN consistently with $n = 5$ qubits, and evaluate the performance on $50$ different images with metrics reported in Table~\ref{comp_QIREN}. 
Results demonstrate that QVF with Siren outperforms QIREN by $30\%$ on MSE. 

Next, we perform representation learning on image collections via latent variable conditioning, i.e., we configure QVF to learn the $50$ images simultaneously. 
Note that QIREN does not support this experimental setting and, hence, we compare QVF with widely used classical foundational INR methods, i.e.,~MLPs with ReLUs, and Siren. 
The comparisons follow the same evaluation protocol, where the only difference between the baselines and QVF is the presence of the ansatz; the results are summarised in Table~\ref{comp_classical}. 
Reconstructed images conditioned on different latent codes are visualised in Fig.~\ref{more_dogs}.

We next consider deployment of QVF on future quantum hardware that could reduce the representation fidelity, such as measurement uncertainty. 
The visual field extracted from the ansatz can differ in its fidelity (accuracy and quality) due to stochastic effects induced by finite sampling on quantum hardware. 
In App.~\ref{s:image_represent_finite_samples}, we visualise in Fig.~\ref{samples} extracted image representations encoded within our pre-trained QVF across a varying number of shots $N_{\text{shot}}$, showing the characteristics of the resulting fields with increasing sampling precision. 

\noindent \textbf{Ansatz Configuration}. 
We next evaluate architectural variations in QVF. 
We investigate the impact of the key hyperparameters: 1) ansatz width, i.e.,~number of qubits $n$; 2) circuit depth $\mathcal{J}$; and 3) latent space dimension $p$ of our classical module \begin{wrapfigure}{r}{0.75\textwidth}
    \vspace{-10pt}
    \includegraphics[width=0.75\textwidth]{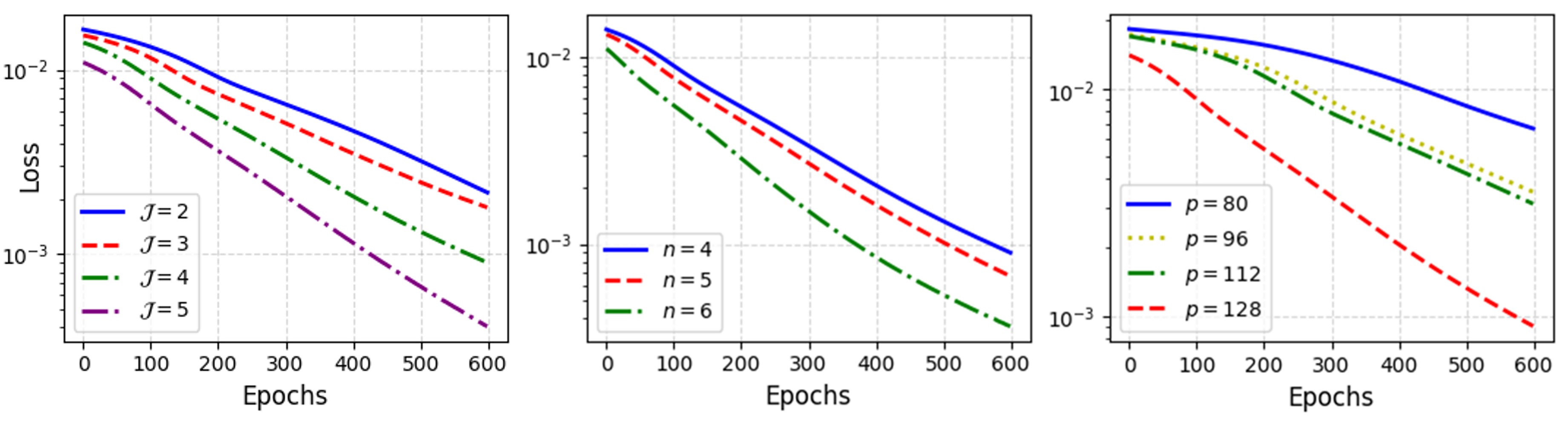}
    \caption{Ablation study with modules influencing the model performance. From left to right: 1) circuit depth $\mathcal{J}$; 2) number of qubits $n$; 3) hidden neuron dimension per layer $p$. 
    }
    \label{ablation}
    \vspace{-10pt}
\end{wrapfigure}for encoding data into quantum states;
see Fig.~\ref{ablation} for the results.
Scaling up the QVF ansatz, i.e.,~increasing $\mathcal{J}$ and width $n$ (while maintaining other parameters), leads to performance gains in both cases. 
QVF scales robustly and is not affected by substantial trainability problems, at least in our evaluated scenario. 
The expressivity of the classical module for integrating non-linear data priors and preparing encoded quantum states serves as the architectural cornerstone. 
With increasing $p$, we observe consistent performance surges. 

\noindent \textbf{Parameter Scaling Analysis}. 
Per design, QVF consists of: 
1) the classical module for neural amplitude encoding; 
and 2) the ansatz. 
As we leverage a tiny MLP in QVF, its parametrisation scales quadratically, i.e.,~$O(p^2)$ w.r.t.~the latent space dimension $p$. 
Meanwhile, our quantum ansatz has parameter scaling of $O(n \mathcal{J})$ w.r.t.~its depth. 
While the ansatz contributes negligibly to the total parameter count, it improves the overall performance by a large margin (see~Fig.~\ref{dog_pic}).  

\subsection{3D (Shape) Representation Learning} 
\label{ssec:3D_comparisons} 

\begin{wrapfigure}{r}{0.48\textwidth} 
    \vspace{-10pt}
    \begin{minipage}{0.48\textwidth}
        \centering
        \scalebox{0.8}{
            \begin{tabular}{@{}l|cc@{}}
                \toprule
                Method & MSE $(\times 10^{-3})$ \textcolor{green}{$\downarrow$} & PSNR \textcolor{red}{$\uparrow$} \\ \midrule
                Ours(Gaussian)+ReLU & 0.98 {\footnotesize$\pm0.09$} & 30.06 {\footnotesize$\pm1.0$} \\
                Ours(Identity)+ReLU & 0.99 {\footnotesize$\pm0.09$} & 30.02 {\footnotesize$\pm1.0$} \\ \midrule
                Ours(Gaussian)+Sin & 0.55 {\footnotesize$\pm0.04$} & 32.59 {\footnotesize$\pm0.2$} \\
                Ours(Identity)+Sin & \textbf{0.54} {\footnotesize$\pm0.04$} & \textbf{32.67} {\footnotesize$\pm0.3$} \\ \midrule
                QIREN~\cite{zhao2024quantum} & 0.78 {\footnotesize$\pm0.05$} & 31.03 {\footnotesize$\pm0.2$} \\ \bottomrule
            \end{tabular}
        }
        \captionof{table}{Numerical results for 2D representation learning between the previous QINR method QIREN~\cite{zhao2024quantum} and our QVF.}
        \vspace{3pt}
        \label{comp_QIREN}
    \end{minipage}
    \hfill
    \begin{minipage}{0.48\textwidth}
        \centering
        \scalebox{0.58}{
            \begin{tabular}{@{}l|cc|c@{}}
                \toprule
                \multirow{2}{*}{Method} & \multicolumn{2}{c|}{Images} & 3D Shapes \\
                & MSE $(\times 10^{-3})$ \textcolor{green}{$\downarrow$} & PSNR \textcolor{red}{$\uparrow$} & MAE $(\times 10^{-3})$ \textcolor{green}{$\downarrow$} \\ \midrule
                Ours(Gaussian)+ReLU & \textbf{1.02} {\footnotesize$\pm0.11$} & \textbf{29.8}{\footnotesize$\pm1.1$} & \textbf{0.99}{\footnotesize$\pm0.07$} \\
                Ours(Identity)+ReLU & 1.03 {\footnotesize$\pm0.09$} & 29.5{\footnotesize$\pm1.2$} & 1.10{\footnotesize$\pm0.09$} \\
                MLP+ReLU \cite{park2019deepsdf} & 2.17 {\footnotesize$\pm0.13$} & 26.57{\footnotesize$\pm0.51$} & 1.43{\footnotesize$\pm0.14$} \\ \midrule
                Ours(Gaussian)+Sin & \textbf{0.62}{\footnotesize$\pm0.05$} & \textbf{32.2}{\footnotesize$\pm0.3$} & \textbf{0.27}{\footnotesize$\pm0.05$} \\
                Ours(Identity)+Sin & 0.72 {\footnotesize$\pm0.06$} & 31.4{\footnotesize$\pm$0.4} & 0.32{\footnotesize$\pm$0.05} \\
                MLP+Sin \cite{Sitzmann2020} & 1.19{\footnotesize$\pm0.08$} & 29.2{\footnotesize$\pm0.4$} & 0.48{\footnotesize$\pm0.06$} \\ \bottomrule
            \end{tabular}
        }
        \captionof{table}{Numerical results for 2D/3D representation learning for our QVF and classical baselines.}
        \label{comp_classical}
    \end{minipage}%
    \vspace{-12pt}
\end{wrapfigure} We next evaluate geometric representation learning of 3D shapes in the form of SDFs. 
This setting is investigated for the first time in the context of QINRs; it poses challenges primarily concerning QINR scalability and the complications arising from varying shape topological structures.
Similar to images, we perform representation learning on 3D shape collections.
We select three shapes from ShapeNet \cite{shapenet2015} and non-uniformly sample signed distances at 100$k$ spatial points per shape, with higher near-surface sampling density for better surface detail capture. 
Note that while the scale of this experiment setup can be considered moderate for classical models, it significantly advances the feasible scale of QINR models (which are nevertheless constrained by the simulator's performance) and provides valuable insights for future advancements. 
We inherit the experimental setting from the experiments with 2D images and report the representational accuracy upon convergence in Table~\ref{comp_classical}.
The baseline setups are, likewise, 
MLPs with different activation functions, including the ReLU, 
which corresponds to the DeepSDF approach~\cite{park2019deepsdf}. 
The final meshes can be extracted from the queried signed distances of QVF using Marching Cubes \cite{lorensen1998marching}, as visualised in Fig.~\ref{chairs}-(a). 
\subsection{Applications Supported by QVF} 
QVF supports applications such as visual field interpolation in the latent space, image inpainting and shape completion. 
Fig.~\ref{Preview}-(c) visualises linear latent-space interpolation of 3D shapes encoded in the converged QVF, i.e.,~a frequent experimental setting in the classical INR literature \cite{park2019deepsdf}. 
QVF also supports image and shape completion by first sampling $\hat{\boldsymbol{z}}$ and optimising its value using MAP; see details in Sec.~\ref{s:training}. 
In the second step, the completion can be performed by leveraging the optimised latent code and inferring missing regions; see Fig.~\ref{chairs}-(b) and App.~\ref{image_inpainting} for the qualitative results. 

\section{Discussion, Future Work and Conclusion} 

Our QVF is a novel QML framework for implicit representation learning of visual fields. 
In our experiments on a quantum hardware simulator, we observe that QVF---even with minimal classical components---can achieve high representation fidelity across data modalities such as images and 3D shapes. 
Furthermore, QVF outperforms the previous QINR approach QIREN (of a similar model scale) and, additionally, is competitive against foundational classical baselines. 
The ansatz configuration and ablation studies highlight the influence of each QVF module.   
Our ablative study confirms sufficient circuit depth, resulting in a balance between the ansatz depth and high representational accuracy. 
Upon our expectations and the theoretical predictions, our QVF is efficient in learning high-frequency signal details (Fig.~\ref{dog_pic}).
As the first among QINR methods, QVF supports joint representation learning of image and 3D shape collections, and applications such as image inpainting and 3D shape completion. 
Finally, we emphasise that this work focuses on the challenges of advancing QINR through fundamental methodological innovations. 
Hence, we do not aim to challenge classical well-engineered models in absolute terms. 
Our implementation can be found on the project page. 
\noindent \textbf{Limitations.} While QVF demonstrates substantial improvements over prior QINR methods in terms of both performance and supported size of visual fields, the current experimental scale, nevertheless, remains constrained due to the quantum hardware simulation overheads. 
Those, however, affect all existing applied QML works 
before the advent of fault-tolerant gate-based quantum computers. 

\noindent \textbf{Future Work.} 
We see various promising avenues for follow-ups and QVF improvements. 
One direction is to explore the preparation of learnable quantum states following a Gibbs distribution with reduced computational complexity (e.g.,~tensor train decomposition \cite{melnikov2023quantum}). 
We also foresee that other problems with open challenges, such as 3D reconstruction and neural rendering from 2D images, could adopt QVF as a representation. 
We also believe that many tricks and further ideas from the INR literature could be adopted in the QINR context in future (e.g.,~space partitioning structures and non-rigid generalisations) \cite{Reiser2021ICCV, tretschk2021nonrigid, Takikawa2021, tewari2022advances}.
\vspace{3pt} 
\noindent \textbf{Author Contributions.} 
SW: Method implementation, experiments, refinement of the concept, draft writing and editing, visualisations and the video;  CT: lab environment, draft editing; VG: method conceptualisation, project coordination, supervision, draft writing and editing. 
\vspace{3pt} 
\noindent \textbf{Acknowledgements.} 
We thank Natacha Kuete Meli, Daniele Lizzio Bosco and Thomas Leimkuehler for helpful comments on the manuscript. 
The work was partially supported by the Deutsche
Forschungsgemeinschaft (DFG, German Research Foundation), project number 534951134.

\small
\bibliographystyle{splncs04}
\bibliography{main}

@String(CVPR= {IEEE Conf. Comput. Vis. Pattern Recog.})

@String(ICCV= {Int. Conf. Comput. Vis.})

@String(BMVC= {Brit. Mach. Vis. Conf.})

@String(CVPR  = {CVPR})

@String(ICCV  = {ICCV})

@String(BMVC  =	{BMVC})

@misc{JWST,
  author = {Jonathan P. Gardner},
  title = {{James Webb Space Telescope}},
  howpublished = "\url{https://webb.nasa.gov/content/multimedia/images.html}",
  year = {2022}, 
  
}

@article{melnikov2023quantum,
  title={Quantum state preparation using tensor networks},
  author={Melnikov, Ar and Termanova, Alena and Dolgov, Sergey and Neukart, Florian and Perelshtein, MR},
  journal={Quantum Science and Technology},
  volume={8},
  number={3},
  pages={035027},
  year={2023},
  publisher={IOP Publishing}
}

@article{jospin2022hands,
  title={Hands-on Bayesian neural networks—A tutorial for deep learning users},
  author={Jospin, Laurent Valentin and Laga, Hamid and Boussaid, Farid and Buntine, Wray and Bennamoun, Mohammed},
  journal={IEEE Computational Intelligence Magazine},
  volume={17},
  number={2},
  pages={29--48},
  year={2022},
  publisher={IEEE}
}

@article{schuld2015introduction,
  title={An introduction to quantum machine learning},
  author={Schuld, Maria and Sinayskiy, Ilya and Petruccione, Francesco},
  journal={Contemporary Physics},
  volume={56},
  number={2},
  pages={172--185},
  year={2015},
  publisher={Taylor \& Francis}
}

@inproceedings{rahaman2019spectral,
  title={On the spectral bias of neural networks},
  author={Rahaman, Nasim and Baratin, Aristide and Arpit, Devansh and Draxler, Felix and Lin, Min and Hamprecht, Fred and Bengio, Yoshua and Courville, Aaron},
  booktitle={International conference on machine learning},
  pages={5301--5310},
  year={2019},
  organization={PMLR}
}

@inproceedings{schleich2024quantum,
  title={Quantum deep equilibrium models},
  author={Schleich, Philipp and Skreta, Marta and Kristensen, Lasse and Vargas-Hernandez, Rodrigo and Aspuru-Guzik, Al{\'a}n},
  booktitle={Advances in Neural Information Processing Systems (NeurIPS)},
  volume={37},
  pages={31940--31967},
  year={2024}
}

@inproceedings{park2019deepsdf,
  title={Deepsdf: Learning continuous signed distance functions for shape representation},
  author={Park, Jeong Joon and Florence, Peter and Straub, Julian and Newcombe, Richard and Lovegrove, Steven},
  booktitle={Computer Vision and Pattern Recognition (CVPR)},
  year={2019}
}

@article{cerezo2021cost,
  title={Cost function dependent barren plateaus in shallow parametrized quantum circuits},
  author={Cerezo, Marco and Sone, Akira and Volkoff, Tyler and Cincio, Lukasz and Coles, Patrick J},
  journal={Nature communications},
  volume={12},
  number={1},
  pages={1791},
  year={2021},
  publisher={Nature Publishing Group UK London}
}

@article{wierichs2022general,
  title={General parameter-shift rules for quantum gradients},
  author={Wierichs, David and Izaac, Josh and Wang, Cody and Lin, Cedric Yen-Yu},
  journal={Quantum},
  volume={6},
  pages={677},
  year={2022}
}

@article{thanasilp2023subtleties,
  title={Subtleties in the trainability of quantum machine learning models},
  author={Thanasilp, Supanut and Wang, Samson and Nghiem, Nhat Anh and Coles, Patrick and Cerezo, Marco},
  journal={Quantum Machine Intelligence},
  volume={5},
  number={1},
  pages={21},
  year={2023},
  publisher={Springer}
}

@article{wang2021noise,
  title={Noise-induced barren plateaus in variational quantum algorithms},
  author={Wang, Samson and Fontana, Enrico and Cerezo, Marco and Sharma, Kunal and Sone, Akira and Cincio, Lukasz and Coles, Patrick J},
  journal={Nature communications},
  volume={12},
  number={1},
  pages={6961},
  year={2021},
  publisher={Nature Publishing Group UK London}
}

@article{holmes2022connecting,
  title={Connecting ansatz expressibility to gradient magnitudes and barren plateaus},
  author={Holmes, Zo{\"e} and Sharma, Kunal and Cerezo, Marco and Coles, Patrick J},
  journal={PRX Quantum},
  volume={3},
  number={1},
  pages={010313},
  year={2022},
  publisher={APS}
}

@inproceedings{weigold2020data,
  title={Data encoding patterns for quantum computing},
  author={Weigold, Manuela and Barzen, Johanna and Leymann, Frank and Salm, Marie},
  booktitle={Conference on Pattern Languages of Programs},
  year={2020}
}

@article{schalkers2024importance,
  title={On the importance of data encoding in quantum Boltzmann methods},
  author={Schalkers, Merel A and M{\"o}ller, Matthias},
  journal={Quantum Information Processing},
  volume={23},
  number={1},
  pages={20},
  year={2024},
  publisher={Springer}
}

@article{bondarenko2020quantum,
  title={Quantum autoencoders to denoise quantum data},
  author={Bondarenko, Dmytro and Feldmann, Polina},
  journal={Physical review letters},
  volume={124},
  number={13},
  pages={130502},
  year={2020},
  publisher={APS}
}

@article{mildenhall2021nerf,
  title={Nerf: Representing scenes as neural radiance fields for view synthesis},
  author={Mildenhall, Ben and Srinivasan, Pratul P and Tancik, Matthew and Barron, Jonathan T and Ramamoorthi, Ravi and Ng, Ren},
  journal={Communications of the ACM},
  volume={65},
  number={1},
  pages={99--106},
  year={2021},
  publisher={ACM New York, NY, USA}
}

@inproceedings{golyanik2020quantum,
  title={A quantum computational approach to correspondence problems on point sets},
  author={Golyanik, Vladislav and Theobalt, Christian},
  booktitle={Computer Vision and Pattern Recognition (CVPR)},
  year={2020}
}

@inproceedings{birdal2021quantum,
  title={Quantum permutation synchronization},
  author={Birdal, Tolga and Golyanik, Vladislav and Theobalt, Christian and Guibas, Leonidas J},
  booktitle={Computer Vision and Pattern Recognition (CVPR)},
  year={2021}
}

@inproceedings{zaech2022adiabatic,
  title={Adiabatic quantum computing for multi object tracking},
  author={Zaech, Jan-Nico and Liniger, Alexander and Danelljan, Martin and Dai, Dengxin and Van Gool, Luc},
  booktitle={Computer Vision and Pattern Recognition (CVPR)},
  pages={8811--8822},
  year={2022}
}

@inproceedings{benkner2021q,
  title={Q-match: Iterative shape matching via quantum annealing},
  author={Benkner, Marcel Seelbach and L{\"a}hner, Zorah and Golyanik, Vladislav and Wunderlich, Christof and Theobalt, Christian and Moeller, Michael},
  booktitle={International Conference on Computer Vision (ICCV)},
  pages={7586--7596},
  year={2021}
}

@article{amin2018quantum,
  title={Quantum boltzmann machine},
  author={Amin, Mohammad H and Andriyash, Evgeny and Rolfe, Jason and Kulchytskyy, Bohdan and Melko, Roger},
  journal={Physical Review X},
  volume={8},
  number={2},
  pages={021050},
  year={2018},
  publisher={APS}
}

@InProceedings{rathi20233d, 
    author={Rathi, Lakshika  and Tretschk, Edith and Theobalt, Christian and Dabral, Rishabh  and Golyanik, Vladislav}, 
    title={{3D-QAE}: Fully Quantum Auto-Encoding of 3D Point Clouds}, 
    booktitle={The British Machine Vision Conference (BMVC)}, 
    year={2023} 
}

@article{bergholm2018pennylane,
  title={Pennylane: Automatic differentiation of hybrid quantum-classical computations},
  author={Bergholm, Ville and Izaac, Josh and Schuld, Maria and Gogolin, Christian and Ahmed, Shahnawaz and Ajith, Vishnu and Alam, M Sohaib and Alonso-Linaje, Guillermo and AkashNarayanan, B and Asadi, Ali and others},
  journal={arXiv preprint 1811.04968},
  year={2018}
}

@article{schuld2021effect,
  title={Effect of data encoding on the expressive power of variational quantum-machine-learning models},
  author={Schuld, Maria and Sweke, Ryan and Meyer, Johannes Jakob},
  journal={Physical Review A},
  volume={103},
  number={3},
  pages={032430},
  year={2021},
  publisher={APS}
}

@article{mcclean2018barren,
  title={Barren plateaus in quantum neural network training landscapes},
  author={McClean, Jarrod R and Boixo, Sergio and Smelyanskiy, Vadim N and Babbush, Ryan and Neven, Hartmut},
  journal={Nature communications},
  volume={9},
  number={1},
  pages={4812},
  year={2018},
  publisher={Nature Publishing Group UK London}
}

@article{zhang2022escaping,
  title={Escaping from the barren plateau via gaussian initializations in deep variational quantum circuits},
  author={Zhang, Kaining and Liu, Liu and Hsieh, Min-Hsiu and Tao, Dacheng},
  journal={Advances in Neural Information Processing Systems},
  volume={35},
  pages={18612--18627},
  year={2022}
}

@article{mitarai2018quantum,
  title={Quantum circuit learning},
  author={Mitarai, Kosuke and Negoro, Makoto and Kitagawa, Masahiro and Fujii, Keisuke},
  journal={Physical Review A},
  volume={98},
  number={3},
  pages={032309},
  year={2018},
  publisher={APS}
}

@inproceedings{farina2023quantum,
  title={Quantum Multi-Model Fitting},
  author={Farina, Matteo and Magri, Luca and Menapace, Willi and Ricci, Elisa and Golyanik, Vladislav and Arrigoni, Federica},
  booktitle={Computer Vision and Pattern Recognition (CVPR)},
  pages={13640--13649},
  year={2023}
}

@article{grant2019initialization,
  title={An initialization strategy for addressing barren plateaus in parametrized quantum circuits},
  author={Grant, Edward and Wossnig, Leonard and Ostaszewski, Mateusz and Benedetti, Marcello},
  journal={Quantum},
  volume={3},
  pages={214},
  year={2019},
  publisher={Verein zur F{\"o}rderung des Open Access Publizierens in den Quantenwissenschaften}
}

@article{kingma2014adam,
  title={Adam: A method for stochastic optimization},
  author={Kingma, Diederik P and Ba, Jimmy},
  journal={arXiv preprint 1412.6980},
  year={2014}
}

@article{cong2019quantum,
  title={Quantum convolutional neural networks},
  author={Cong, Iris and Choi, Soonwon and Lukin, Mikhail D},
  journal={Nature Physics},
  volume={15},
  number={12},
  pages={1273--1278},
  year={2019},
  publisher={Nature Publishing Group UK London}
}

@inproceedings{xie2022neural,
  title={Neural fields in visual computing and beyond},
  author={Xie, Yiheng and Takikawa, Towaki and Saito, Shunsuke and Litany, Or and Yan, Shiqin and Khan, Numair and Tombari, Federico and Tompkin, James and Sitzmann, Vincent and Sridhar, Srinath},
  booktitle={Computer Graphics Forum},
  volume={41},
  pages={641--676},
  year={2022},
  organization={Wiley Online Library}
}

@article{shapenet2015,
  title       = {{ShapeNet: An Information-Rich 3D Model Repository}},
  author      = {Chang, Angel X. and Funkhouser, Thomas and Guibas, Leonidas and Hanrahan, Pat and Huang, Qixing and Li, Zimo and Savarese, Silvio and Savva, Manolis and Song, Shuran and Su, Hao and Xiao, Jianxiong and Yi, Li and Yu, Fisher},
  journal      = {arXiv preprint 1512.03012},
  year        = {2015}
}

@inproceedings{chen2017stylebank,
  title={Stylebank: An explicit representation for neural image style transfer},
  author={Chen, Dongdong and Yuan, Lu and Liao, Jing and Yu, Nenghai and Hua, Gang},
  booktitle={Computer Vision and Pattern Recognition (CVPR)},
  pages={1897--1906},
  year={2017}
}

@article{kerenidis2019q,
  title={q-means: A quantum algorithm for unsupervised machine learning},
  author={Kerenidis, Iordanis and Landman, Jonas and Luongo, Alessandro and Prakash, Anupam},
  journal={Advances in Neural Information Processing Systems (NeurIPS)},
  volume={32},
  year={2019}
}

@article{rebentrost2014quantum,
  title={Quantum support vector machine for big data classification},
  author={Rebentrost, Patrick and Mohseni, Masoud and Lloyd, Seth},
  journal={Physical review letters},
  volume={113},
  number={13},
  pages={130503},
  year={2014},
  publisher={APS}
}

@article{lloyd2014quantum,
  title={Quantum principal component analysis},
  author={Lloyd, Seth and Mohseni, Masoud and Rebentrost, Patrick},
  journal={Nature Physics},
  volume={10},
  number={9},
  pages={631--633},
  year={2014},
  publisher={Nature Publishing Group UK London}
}

@inproceedings{li20223d,
  title={3d neural scene representations for visuomotor control},
  author={Li, Yunzhu and Li, Shuang and Sitzmann, Vincent and Agrawal, Pulkit and Torralba, Antonio},
  booktitle={Conference on Robot Learning},
  pages={112--123},
  year={2022},
  organization={PMLR}
}

@inproceedings{molaei2023implicit,
  title={Implicit neural representation in medical imaging: A comparative survey},
  author={Molaei, Amirali and Aminimehr, Amirhossein and Tavakoli, Armin and Kazerouni, Amirhossein and Azad, Bobby and Azad, Reza and Merhof, Dorit},
  booktitle={International Conference on Computer Vision (ICCV)},
  pages={2381--2391},
  year={2023}
}

@inproceedings{tschernezki2022neural,
  title={Neural Feature Fusion Fields: 3D distillation of self-supervised 2D image representations},
  author={Tschernezki, Vadim and Laina, Iro and Larlus, Diane and Vedaldi, Andrea},
  booktitle={International Conference on 3D Vision (3DV)},
  pages={443--453},
  year={2022} 
}

@article{schuld2020circuit,
  title={Circuit-centric quantum classifiers},
  author={Schuld, Maria and Bocharov, Alex and Svore, Krysta M and Wiebe, Nathan},
  journal={Physical Review A},
  volume={101},
  number={3},
  pages={032308},
  year={2020},
  publisher={APS}
}

@inproceedings{zhao2024quantum, 
  title={Quantum implicit neural representations}, 
  author={Zhao, Jiaming and Qiao, Wenbo and Zhang, Peng and Gao, Hui}, 
  booktitle={International Conference on Machine Learning (ICML)}, 
  year={2024} 
}

@article{ball2023boltzmann,
  title={Boltzmann distributions on a quantum computer via active cooling},
  author={Ball, Carter and Cohen, Thomas D},
  journal={Nuclear Physics A},
  volume={1038},
  pages={122708},
  year={2023},
  publisher={Elsevier}
}

@incollection{lorensen1998marching,
  title={Marching cubes: A high resolution 3D surface construction algorithm},
  author={Lorensen, William E and Cline, Harvey E},
  booktitle={Seminal graphics: pioneering efforts that shaped the field},
  pages={347--353},
  year={1998},
  publisher = {ACM}
}

@article{krizhevsky2009learning,
  title={Learning multiple layers of features from tiny images},
  author={Krizhevsky, Alex and Hinton, Geoffrey and others},
  year={2009},
  journal={University of Toronto}
}

@inproceedings{tewari2022advances,
  title={Advances in neural rendering},
  author={Tewari, Ayush and Thies, Justus and Mildenhall, Ben and Srinivasan, Pratul and Tretschk, Edgar and Yifan, Wang and Lassner, Christoph and Sitzmann, Vincent and Martin-Brualla, Ricardo and Lombardi, Stephen and others},
  booktitle={Computer Graphics Forum},
  volume={41},
  pages={703--735},
  year={2022},
  organization={Wiley Online Library}
}

@article{shiba2019convolution,
  title={Convolution filter embedded quantum gate autoencoder},
  author={Shiba, Kodai and Sakamoto, Katsuyoshi and Yamaguchi, Koichi and Malla, Dinesh Bahadur and Sogabe, Tomah},
  journal={arXiv preprint 1906.01196},
  year={2019}
}

@article{baek20223d,
  title={3D scalable quantum convolutional neural networks for point cloud data processing in classification applications},
  author={Baek, Hankyul and Yun, Won Joon and Kim, Joongheon},
  journal={arXiv preprint 2210.09728},
  year={2022}
}

@inproceedings{meli2025qucoop,
  title={{QuCOOP}: A Versatile Framework for Solving Composite and Binary-Parametrised Problems on Quantum Annealers},
  author={Natacha Kuete Meli and Vladislav Golyanik and Marcel Seelbach Benkner and Michael Moeller},
  booktitle={Computer Vision and Pattern Recognition (CVPR)},
  year={2025}
}

@article{meli2025quantum,
  title={Quantum-enhanced Computer Vision: Going Beyond Classical Algorithms},
  author={Meli, Natacha Kuete and Wang, Shuteng and Benkner, Marcel Seelbach and Sasdelli, Michele and Chin, Tat-Jun and Birdal, Tolga and Moeller, Michael and Golyanik, Vladislav},
  journal={arXiv preprint arXiv:2510.07317},
  year={2025}
}

@article{Benedetti2019, 
year = {2019},
volume = {4},
number = {4},
pages = {043001},
author = {Marcello Benedetti and Erika Lloyd and Stefan Sack and Mattia Fiorentini},
title = {Parameterized quantum circuits as machine learning models},
journal = {Quantum Science and Technology} 
}

@inproceedings{Sitzmann2020, 
 author = {Sitzmann, Vincent and Martel, Julien and Bergman, Alexander and Lindell, David and Wetzstein, Gordon}, 
 booktitle = {Advances in Neural Information Processing Systems (NeurIPS)}, 
 pages = {7462--7473}, 
 title = {Implicit Neural Representations with Periodic Activation Functions}, 
 volume = {33}, 
 year = {2020} 
}

@inproceedings{chen2021learning,
  title={Learning continuous image representation with local implicit image function},
  author={Chen, Yinbo and Liu, Sifei and Wang, Xiaolong},
  booktitle={Computer Vision and Pattern Recognition (CVPR)},
  year={2021}
}

@article{Haar1933, 
 author = {Alfred Haar}, 
 journal = {Annals of Mathematics}, 
 number = {1}, 
 pages = {147--169}, 
 title = {Der Massbegriff in der Theorie der Kontinuierlichen Gruppen}, 
 volume = {34}, 
 year = {1933} 
}

@article{google_willow,
  title={Quantum error correction below the surface code threshold},
  author={{Google Quantum AI and Collaborators}},
  journal={Nature},
  year={2024}
}

@INPROCEEDINGS{Reiser2021ICCV,
          author = {Christian Reiser and Songyou Peng and Yiyi Liao and Andreas Geiger},
          title = {KiloNeRF: Speeding up Neural Radiance Fields with Thousands of Tiny MLPs},
          booktitle = {International Conference on Computer Vision (ICCV)},
          year = {2021}
}

@inproceedings{tretschk2021nonrigid, 
    title = {Non-Rigid Neural Radiance Fields: Reconstruction and Novel View Synthesis of a Dynamic Scene From Monocular Video}, 
    author = {Tretschk, Edgar and Tewari, Ayush and Golyanik, Vladislav and Zollh\"{o}fer, Michael and Lassner, Christoph and Theobalt, Christian}, 
    booktitle = {International Conference on Computer Vision ({ICCV})},
    year = {2021} 
}

@InProceedings{Takikawa2021, 
    author    = {Takikawa, Towaki and Litalien, Joey and Yin, Kangxue and Kreis, Karsten and Loop, Charles and Nowrouzezahrai, Derek and Jacobson, Alec and McGuire, Morgan and Fidler, Sanja}, 
    title     = {Neural Geometric Level of Detail: Real-Time Rendering With Implicit 3D Shapes}, 
    booktitle = {Computer Vision and Pattern Recognition (CVPR)}, 
    year      = {2021} 
}

\clearpage
\appendix

\begin{center}
\Large{\textbf{Supplementary Material}}
\vspace{20pt}
\end{center}

This appendix supplements the main paper, starting in Sec.~\ref{s:background} with a detailed background on gate-based quantum computing. 
It covers both quantum physics foundations and applications of quantum computing in machine learning and INR. 
It also connects quantum circuit measurements and Bayesian inference. 
Next, we outline the full algorithmic (training) protocol; visualise the quantum ansatz architecture used in the experiments; and provide more implementation details in Sec.~\ref{alg_protocol}. 
On the experimental side, we further analyse QVF performance with noisy circuits in Sec.~\ref{gate_noise}, followed by the image representation quality in dependence on the number of measurement repetitions (shots) in Sec.~\ref{s:image_represent_finite_samples}.
Applications supported by QVF, such as image inpainting and shape completion, are discussed in Secs.~\ref{image_inpainting} and \ref{s:Noisy_shape_completion}, while additional visualisations of 3D shapes are shown in Sec.~\ref{additional_3D}.
Sec.~\ref{ss:gate-based hardware} then discusses the development status of currently available real quantum hardware. 

\section{Background} \label{s:background}
\subsection{Preliminaries on Gate-based Quantum Computing}\label{ssec:Preliminaries}

\noindent \begin{wrapfigure}{r}{0.27\textwidth}
\vspace{-15pt}
\includegraphics[width=0.95\linewidth]{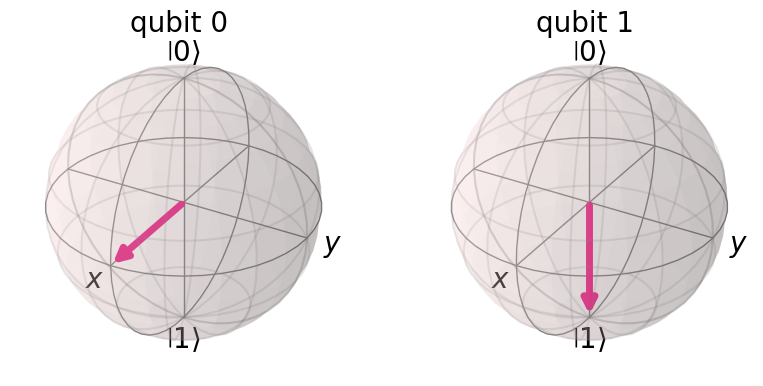}
\caption{Bloch sphere visualisation of qubit states. Qubit 0: $ \ket{\psi} = \frac{1}{\sqrt{2}} (\ket{0} + \ket{1}) $, qubit 1: $ \ket{\psi} =  \ket{1} $.}
\label{Bloch Spheres}
\vspace{-10pt}
\end{wrapfigure} \textbf{Qubits}. The fundamental information blocks of a quantum processing unit~(QPU) are qubits, i.e.,~the analogues of bits in classical computing. 
Unlike classical bits deterministically representing one possible state~(0 or 1), qubits can statistically represent two distinct information states at the same time, denoted in the bra--ket notation as $ \ket{0} $ and $ \ket{1} $.

\textit{Superposition} is a fundamental property distinguishing qubits from bits: It grants qubits the capacity to exist in a combinatorial state $ \ket{\psi} $ of $ \ket{0} $ and $ \ket{1} $ such that:

\begin{equation} \label{eq:13}
\ket{\psi} = \alpha  \ket{0} + \beta \ket{1}, 
\end{equation}

\noindent with $ \alpha, \beta  \in \mathbb{C} $ and $ | \alpha |^2 + | \beta |^2 = 1 $. 
Qubit states $\ket{\psi}$ can be visualised on Bloch spheres (see Fig.~\ref{Bloch Spheres}) or expressed in a vector form: 
\begin{equation} \label{eq:14}
 \ket{0} = \begin{bmatrix}1 \\ 0\end{bmatrix},  \ket{1} = \begin{bmatrix}0 \\ 1\end{bmatrix}, \ket{\psi} = \alpha  \ket{0} + \beta \ket{1} = \begin{bmatrix}\alpha \\ \beta\end{bmatrix}.
\end{equation}

\noindent \textit{Measurement} in quantum mechanics inherently adopts a statistical approach to extract numerical information. For a qubit state $ \ket{\psi} = \alpha \ket{0} + \beta \ket{1} $ measured with operator $\hat{O}$~(that must be Hermitian, i.e.,~$\hat{O}^{\dagger} = \hat{O}$), this implies probabilities $ |\alpha|^2 $ and $ |\beta|^2 $, respectively, for measuring the information~(i.e.,~eigenvalue of the measurement operator $\hat{O}$) stored in states $ \ket{0} $ and $ \ket{1} $: 
\begin{equation} 
    \hat{O} \ket{0} = \kappa \ket{0} \;\text{and} \; \hat{O} \ket{1} = \delta \ket{1}, 
\end{equation} 
where $\kappa$ and $\delta$ are 
eigenvalues of the measurement operator $\ket{O}$. 
The key aspect of measurement is the phenomenon known as wave function collapse, i.e.,~the projective measurement causes $ \ket{\psi} $ to collapse to the operator's eigenstate, $\ket{0}$ or $\ket{1}$, conditioned on the measurement, i.e.,~$\kappa$ or $\delta$. 
\textit{Entanglement} further distinguishes quantum from classical computing. In the classical case, information stored in bits is independent, i.e.,~measuring one bit does not affect others. 
In the quantum realm, qubits can be highly correlated, exhibiting entanglement such that the information of one qubit can be interrelated with another, despite possible physical distance between them. 
For instance, a general information state of a 2-qubit system $\ket{\psi}_2$ can be expressed as: 
\begin{equation} \label{eq:15}
\ket{\psi}_2 = a \ket{00} + b \ket{01} + c \ket{10} + d \ket{11}, 
\end{equation}
with $ a, b, c, d \in \mathbb{C} $ such that $ |a|^2 + |b|^2 + |c|^2 + |d|^2 = 1 $. The 2-qubit system is considered entangled if $ \ket{\psi}_2 $ cannot be expressed as a tensor product of two qubits $ \ket{\psi}_{a1} $ and $ \ket{\psi}_{a2} $, indicating that their information cannot be independently measured without disturbing each other, i.e.,
\begin{equation} \label{eq:16}
\ket{\psi}_2 \neq \ket{\psi}_{a1} \otimes \ket{\psi}_{a2}. 
\end{equation}
\noindent\textbf{Rotation Operators}. The operators responsible for rotating quantum states $\ket{\psi}$ of qubits along $x, y, z$ axes on a Bloch sphere are referred to as rotation operators. Any single qubit operator $\hat{R}$ can be expressed as a combination of such rotation operators $ \hat{R_x}, \hat{R_y}, \hat{R_z} $, i.e.,~$\hat{R}(\theta,\tau,\gamma) = \hat{R_x}(\theta)\hat{R_y}(\tau)\hat{R_z}(\gamma) $ with angles $\theta, \tau$ and $\gamma$: 
\begin{equation} \label{eq:17}
\hat{R_x} (\theta) = \begin{bmatrix} \cos(\frac{\theta}{2}) & -i \sin(\frac{\theta}{2})\\ -i \sin(\frac{\theta}{2}) & \cos(\frac{\theta}{2})\end{bmatrix}, \quad \hat{R_y} (\tau) = \begin{bmatrix}\cos(\frac{\tau}{2}) & - \sin(\frac{\tau}{2})\\ \sin(\frac{\tau}{2}) & \cos(\frac{\tau}{2})\end{bmatrix}, \quad \hat{R_z} (\gamma) = \begin{bmatrix} e^{-i \frac{\gamma}{2}} & 0\\ 0 & e^{i \frac{\gamma}{2}}\end{bmatrix}.
\end{equation}
The Pauli operators $ {\hat{X}, \hat{Y}, \hat{Z}} $ represent specific instances of above rotation operators, i.e.~rotations by $\pi$ radians along the $x, y, z$-axes, respectively. 
These operators can also be expressed as matrices in the computational basis $ \ket{0}, \ket{1} $ as follows: 
\begin{equation} \label{eq:20}
\hat{X} = \begin{bmatrix}0 & 1\\ 1 & 0\end{bmatrix}, \hat{Y} = \begin{bmatrix}0 & -i\\ i & 0\end{bmatrix}, \hat{Z} = \begin{bmatrix}1 & 0\\ 0 & -1\end{bmatrix}. 
\end{equation}
\noindent\textbf{The Schrödinger Equation}. Quantum computing involves the manipulation of information according to the principles of quantum mechanics, with its foundation rooted in the time-dependent Schrödinger equation: 
\begin{equation} \label{eq:21}
i \hbar \frac{d}{dt} \ket{\psi(t)} = \hat{H}(t) \ket{\psi(t)}, 
\end{equation}
where $\hbar$ is Planck's constant, and $\ket{\psi(t)}$ and $\ket{\psi(0)}$ are the quantum states after and before evolution, respectively; 
$\hat{H}$ is the Hamiltonian operator of the quantum system. 
Therefore, the evolution of quantum states can be described by the following relationship: 

\begin{equation} \label{eq:22}
    \ket{\psi(t)} = \hat{T} e^{-\frac{i}{\hbar} \int_{0}^{t} \hat{H}(t)dt } \ket{\psi(0)}, 
\end{equation}
with $\hat{T}$ denoting the time ordering operator. 
This simplifies to $ e^{-\frac{it}{\hbar} \hat{H} } \ket{\psi(0)}$ for time-independent $ \hat{H} $. Using a more compact notation, the Schrödinger equation can also be equivalently written as: 
\begin{equation} \label{eq:23}
 \ket{\psi(t)} = \hat{U}(\hat{H}, t) \ket{\psi(0)}, \,\text{with}
\end{equation}
\begin{equation} \label{eq:24}
 \space \hat{U}(\hat{H}, t) = e^{-\frac{it}{\hbar} \hat{H} }. 
\end{equation}
To perform rotation operations on qubits, the system Hamiltonian $ \hat{H} $ can be set to $ E \hat{\sigma}, $ with $ \hat{\sigma} \in \{ \hat{X}, \hat{Y}, \hat{Z} \}$. By substituting $ \eta = 2Et/ \hbar $, we arrive at:
\begin{equation}
    \hat{U}(\hat{H}, t) = e^{-\frac{it}{\hbar} \hat{H} } = e^{-\frac{i \alpha}{2} \hat{\sigma} } = \hat{R_{\sigma}}(\eta). 
\end{equation} 

\subsection{Review: Quantum Machine Learning}\label{ssec:background_appendix} 

\noindent The potential of quantum computing to enhance machine learning algorithms leads to the emergence of quantum machine learning (QML)~\cite{schuld2015introduction}, a discipline employing quantum mechanical phenomena to tackle classically intractable learning problems through enhanced computational paradigms~\cite{schuld2015introduction}.
Central to QML are: 1) a feature map, which encodes classical input data into quantum states and 2) a variational ansatz, which performs quantum transformations on the quantum states.
PQC have been shown to be asymptotic universal function approximators~\cite{Benedetti2019, schuld2021effect}. 
Several standardised QML  algorithms have been explored, including quantum principal component analysis~\cite{lloyd2014quantum}, quantum support vector machines~\cite{rebentrost2014quantum}, quantum Boltzmann machines~\cite{amin2018quantum}, and quantum k-means clustering~\cite{kerenidis2019q}. 

\noindent \textbf{Feature Map}. Integrating classical Euclidean data $\mathbf{x}$ into quantum computational frameworks necessitates a non-trivial mapping to quantum states $\ket{\psi(\mathbf{x})}$ in a Hilbert space $\mathcal{H}$. 
Several established encoding techniques exist, including \textit{basis encoding}, \textit{amplitude encoding}, \textit{Hamiltonian evolution encoding}, with each presenting distinct trade-offs in qubit efficiency and circuit depth complexity.
However, the determination of optimal encoding schemes remains an open research challenge, as the relationship between encoding fidelity $\mathcal{F}(\mathbf{x}) = |\braket{\psi_{\text{ideal}}(\mathbf{x})|\psi_{\text{encoded}}(\mathbf{x})}|^2$, resource requirements, and task-specific performance metrics (e.g., classification accuracy or function approximation error $\epsilon$) remains poorly characterised across different problem domains.

\noindent \textbf{Variational Ansatz}. Quantum evolution of classical information embedded in states $\ket{\psi(x)}$ requires parameterised unitary ansatz $\hat{U}(\theta) \in \mathbb{C}^{2^n \times 2^n}$ acting on $n$-qubit systems.
Physically, the ansatz is constructed through sequential composition of such unitary transformations, formally expressed as $\hat{U}(\boldsymbol{\theta}) = \mathcal{T}\left(\prod_{i=1}^t \hat{U}_i(\theta_i)\right)$, where $\mathcal{T}$ denotes the time-ordering operator governing gate sequence implementation.
This induces a Hilbert space transformation $\hat{U}(\boldsymbol{\theta}): \mathcal{H} \to \mathcal{H}$ that maps input states to processed output states through the operation $\ket{\phi(x, \boldsymbol{\theta})} = \hat{U}(\boldsymbol{\theta})\ket{\psi(x)}$.

\noindent \textbf{Measurement}. Quantum computation culminates in statistical data extraction from evolved quantum states $\ket{\phi(x)}$ through projective measurements using Hermitian observables $\hat{O}$, where the computational output is formally defined as the expectation value:  $V(x) = \bra{\phi(x)}\hat{O}{\ket{\phi(x)}}$.
Measurements collapse quantum states according to the Born rule, thereby restricting access to the embedded classical information to statistical estimators derived from repeated measurements.
The choice of observable $\hat{O}$ fundamentally governs both the information-theoretic capacity of the measurement protocol and its computational complexity.

\noindent \textbf{Training a Variational Ansatz}. Instead of constructing a computational graph and performing backpropagation, training quantum circuits involves only forward evaluations~\cite{wierichs2022general}.
To minimize a measurement-dependent cost function $\mathcal{L}(\boldsymbol{\theta})$, the exact gradients $\nabla\mathcal{L}(\boldsymbol{\theta})$ can be evaluated through quantum circuit evaluations at shifted parameters $\boldsymbol{\theta} \pm \frac{\pi}{2}\mathbf{e}_i$ for basis vectors $\mathbf{e}_i$, expressed as:
\begin{equation}
    \partial_{\theta_i}\mathcal{L} = \frac{1}{2}\left[\mathcal{L}\left(\theta_i + \frac{\pi}{2}\right) - \mathcal{L}\left(\theta_i - \frac{\pi}{2}\right)\right].
\end{equation}
This technique, i.e. parameter-shift rule, exploits the trigonometric structure of unitary gate generators $\hat{G}_i$ (where $\hat{U}_i(\theta_i) = e^{-i\theta_i\hat{G}_i}$) to enable hardware-compatible gradient estimation without numerical approximation or persistent circuit memory - a critical advantage over classical backpropagation that requires differentiable computational graphs.

\subsection{Review: Barren Plateaus} \label{barren_pleatau}

\noindent Training a variational ansatz $\hat{S}(\boldsymbol{\theta})$ is fundamentally constrained by the \textit{barren plateau} phenomenon, where random parameter initialisation induces exponential vanishing of cost function gradients across Hilbert space. As formally demonstrated by McClean et al.~\cite{mcclean2018barren} through concentration of measure analysis: 

\noindent \textit{``...for a wide class of reasonable parametrised quantum circuits, the probability that the gradient along any reasonable direction is non-zero to some fixed precision is exponentially small as a function of the number of qubits.''} 

This phenomenon is also known as \textit{barren plateau}, which can be expressed mathematically for a system with $n$ qubits as follows: 
\begin{equation} \label{eq:26}
    \mathbb{E}_w [\partial _w L(w)] = 0, \quad \operatorname{Var}_w [\partial _w L(w)] \in O\Big(\frac{1}{\nu^n}\Big), \;\nu > 1,
\end{equation} 
where $\nu$ characterises the circuit's entangling capacity.
The variance bound's scaling establishes that gradient estimators require $\mathcal{O}(\nu^n)$ measurement samples to maintain constant precision, resulting in an exponential resource overhead that renders practical optimisation infeasible for $n \gg 1$. 
This poses challenges, particularly for gradient-based learning. 
Identified factors contributing to barren plateaus
include observable locality~\cite{cerezo2021cost,thanasilp2023subtleties}, specific noise models~\cite{wang2021noise}, and an ansatz close to a 2-design, i.e.,~matching Haar random unitaries up to the second moment~\cite{mcclean2018barren,holmes2022connecting}. 
Those highlight the importance of selecting appropriate initialisation protocols, quantum ansatz designs and observables. 

\subsection{Connection of PQCs to Bayesian Inference} \label{s:Bayseian_inference} 

As quantum circuits are inherently probabilistic models, they share conceptual parallels with Bayesian inference. 
In Bayesian neural networks (BNNs), probabilistic outputs emerge from parameters governed by prior distributions, with training focused on maximising the conditional likelihood of observed data labels while implicitly updating a posterior distribution over the parameters.
For PQC, while they leverage deterministic parameters, they exhibit probabilistic outputs due to the stochastic nature of quantum measurements, which---under sufficiently large shot counts---approximate Gaussian distributions in accordance with the CLT. 
While the probabilistic outputs of PQC permit an interpretative lens rooted in Bayesian principles, their training does not inherently involve posterior inference over parameters unless explicitly cast within a Bayesian formalism \cite{jospin2022hands}.  
This distinction underscores that the Bayesian interpretation of PQCs arises from their measurement statistics rather than an intrinsic probabilistic parameter space. 

\begin{algorithm}[t]
\caption{QVF Training Protocol}
\label{alg:qvf_protocol}
\begin{algorithmic}[1]
    \STATE \textbf{Input:} Training dataset $X = \{(\Theta_i, s_i)\}_{i=1}^W$; number of qubits $n$; epochs $N_{\text{epoch}}$; measurement shots $N_{\text{shot}}$; parameters $\boldsymbol{\theta}$ = $\{\boldsymbol{\theta}_q,\boldsymbol{\theta}_c\}$; inverse temperature $\beta$. 
    
    \FOR{epoch $= 1$ \TO $N_{\text{epoch}}$}
        \STATE \textbf{Classical Inference (Sec.~\ref{ss:energy}):} 
        \STATE \quad Compute energy spectrum $\mathbf{E}(\Theta_i;\boldsymbol{\theta}_c)$.
        \STATE \quad Evaluate Gibbs distribution: 
        \[
            P_i = \frac{e^{-\beta E_i}}{Z}, \; \text{where} \; Z = e^{-\beta E_j}. 
        \]
        
        \STATE \textbf{Quantum State Preparation:}
        \STATE \quad Initialise $\hat{\rho}_0 = \sum_{i=1}^{2^n} P_i \ket{i}\bra{i}$.
        
        \STATE \textbf{Quantum Evolution (Sec.~\ref{s:circuit}):} 
        \STATE \quad Apply ansatz $\hat{S}(\boldsymbol{\theta}_q) = \prod_{\ell=1}^{J} e^{-i\theta_{q,\ell}\hat{H}_{\ell}}$ to obtain:
        \[
            \hat{\rho}(\boldsymbol{\theta}_q) = \hat{S}(\boldsymbol{\theta}_q)\hat{\rho}_0\hat{S}^\dagger(\boldsymbol{\theta}_q).
        \]
        
        \STATE \textbf{Measurement and Observables:}
        \STATE \quad Estimate $\langle \hat{M}_k \rangle = \text{Tr}[\hat{\rho}(\boldsymbol{\theta}_q)\hat{M}_k]$ for $k=1,\dots,K$.
        
        \STATE \textbf{Gradient Computation of the Loss $\mathcal{L}$ (Sec.~\ref{s:training}}): 
        \STATE \quad Quantum: $\partial\langle\hat{M}_k\rangle/\partial\boldsymbol{\theta}_q$ via parameter-shift rule.
        \STATE \quad \parbox[t]{\linewidth}{Classical: $\nabla_{\boldsymbol{\theta}_c}\mathcal{L}$ via automatic differentiation \\ through $\mathbf{E}(\Theta_i;\boldsymbol{\theta}_c)$.}

        \STATE \textbf{Parameter Update:} 
        \STATE \quad Adam optimiser step with learning rate $\eta$:
        \[
            \boldsymbol{\theta} \leftarrow \boldsymbol{\theta} - \eta \left( \nabla_{\boldsymbol{\theta}_c}\mathcal{L}, \nabla_{\boldsymbol{\theta}_q}\mathcal{L} \right). 
        \]
    \ENDFOR
    \STATE \textbf{Output:} Optimised parameters $\boldsymbol{\theta}_q^*$, $\boldsymbol{\theta}_c^*$. 
\end{algorithmic}
\end{algorithm}

\section{Algorithmic Protocol and Ansatz} \label{alg_protocol}

We provide the complete training protocol of QVF in Alg.~\ref{alg:qvf_protocol}. 

\subsection{Detailed Variational Ansatz Visualisation} \label{s:ansatz visualisation}

\begin{figure*}[ht]
    \centering
    \includegraphics[width = 1.0\textwidth]{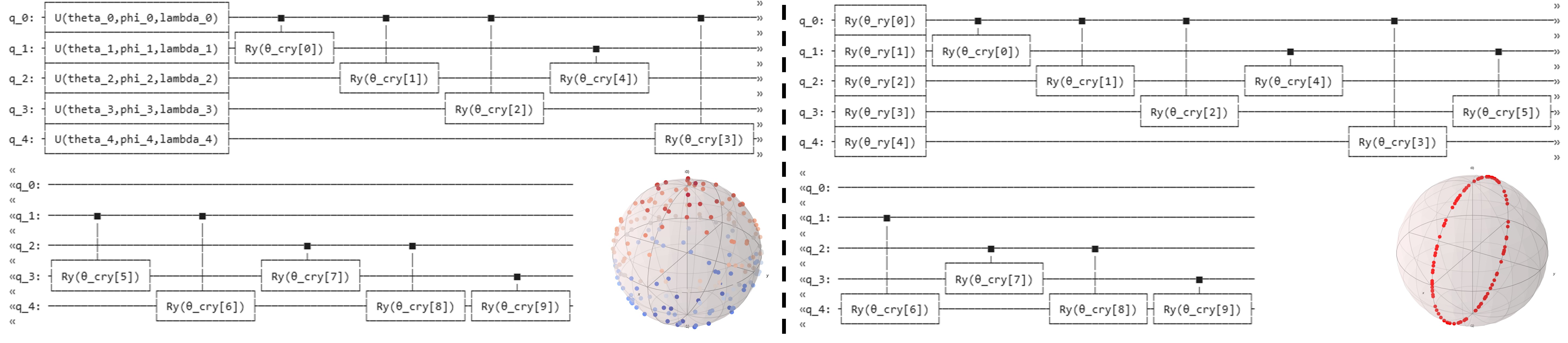} 
    \caption{Visualisation of ansatz designs and their exemplary induced traversable quantum states within Hilbert space: strongly entangled ansatz (left) and QVF (right). 
    Traversable states of different circuit ansatze are visualised on the bottom right for both ansatze.} 
    \label{ansatz_visualisation}
    \vspace{-10pt}
\end{figure*}

We also visualise our QVF ansatz with bounded Haar randomness. 
Fig.~\ref{ansatz_visualisation} compares the reachable Hilbert states between ours and the ansatz in QIREN that does not restrict the set of possible operations to real-valued unitaries~\cite{schuld2020circuit,zhao2024quantum}. 
We highlight the circuit structures via their parametrised single-qubit rotations and inter-qubit entanglement patterns. 
Traversable quantum states are visualised on the Bloch sphere by sampling the ansatz. 

\subsection{Additional Implementation Details} \label{imple_details}

We next provide additional implementation details on our experimental setup. 
\noindent \textbf{Parameterisation of QVF vs.~QIREN.} QVF is designed to be a compact QINR model. Compared to QIREN, QVF does not need classical post-processing while maintaining high representational accuracy. The number of parameters in our experiments is $0.52 \cdot 10^5$ (corresponding to $p = 128$).
vs.~$0.74 \cdot 10^5$ for QVF and QIREN, respectively. 
The ansatz of QVF is configured with $\mathcal{J}{=}5$ and $n{=}5$, and for QIREN, we use the default depth. 
\noindent \textbf{Simulation Stability in Preparing Quantum States following Gibbs distribution.} When simulating the preparation of such quantum states, the partition function evaluation involves exponentiations of large and small Hamiltonian eigenvalues; see Eqs.~\eqref{eq:3} and \eqref{eq:partition_function}, which could cause numerical instabilities. We leverage the log-sum-exp trick, a well-established numerical stabilisation technique for this problem. 

\section{Image Representation with Noisy Circuits} \label{gate_noise}

\begin{wraptable}{r}{0.6\textwidth}
\centering
\vspace{-10pt}
\scalebox{0.75}{
\begin{tabular}{@{}l|cccc@{}}
\toprule
Method                 & no noise & $\sigma$ = 0.01 & $\sigma$ = 0.05  & $\sigma$ = 0.1   \\ \midrule
Ours (Gaussian) + ReLU & 30.06 $\pm$ 0.1    & 30.1 $\pm$ 0.1 & 28.42 $\pm$ 0.1 & 25.78 $\pm$ 0.2 \\
Ours (Identity) + ReLU & 30.02 $\pm$ 0.2    & 29.6 $\pm$ 0.1 & 27.98 $\pm$ 0.2 & 25.96 $\pm$ 0.2 \\
Ours (Gaussian) + Sin  & 32.59 $\pm$ 0.2    & 32.4 $\pm$ 0.2 & 30.66 $\pm$ 0.1 & 27.94 $\pm$ 0.2 \\
Ours (Identity) + Sin  & 32.67 $\pm$ 0.3    & 32.8 $\pm$ 0.2 & 30.34 $\pm$ 0.2 & 28.12 $\pm$ 0.2 \\ \bottomrule
\end{tabular}
} 
\caption{QVF performance with noisy circuits. $\sigma$ is the perturbation ratio modelling quantum circuit infidelities.} 
\vspace{-15pt}
\label{tab:noise_experiments}
\end{wraptable}  Evaluation with quantum circuit noise can provide valuable insights for the practical deployment of QVF on near-term quantum hardware. 
Hence, we investigate the influence of quantum gate infidelities on the performance of QVF, i.e.,~a dominant source of errors and noise in the quantum operations. 
Gate operation infidelity arises from intrinsic control imperfections in quantum hardware, resulting in stochastic deviations of the performed gate operations from their expected behaviour. 
These imperfections constrain gate fidelity to finite precision, which can be effectively modelled as zero-mean perturbations to the gate parameters within a bounded range.
To simulate the impact of such noise, we introduce zero-mean Gaussian perturbations with varying standard deviations: Higher values correspond to the noise levels typical for current near-term quantum devices, while lower values reflect anticipated improvements in future hardware.
The experiments are performed on selected 2D images using the same quantum hardware simulator of PennyLane \cite{bergholm2018pennylane} as in the main matter (Sec.~\ref{ssec:2D_comparisons}). 
We report the results with different levels of quantum gate perturbation ratio $\sigma$ in Table~\ref{tab:noise_experiments} to quantify the degradation in performance under various noise regimes. 
As expected, increasing $\sigma$ leads to a decrease in 2D image PSNR. 
Even with $\sigma = 0.1$, we achieve a PSNR of ${\sim}25$dB or higher (cf.~Fig.~\ref{samples}). 

\section{Image Representation with a Different Number of Samples (Shots)} \label{s:image_represent_finite_samples} 

\noindent \begin{wrapfigure}{r}{0.45\textwidth}
    \vspace{-15pt}
    \includegraphics[width=\linewidth]{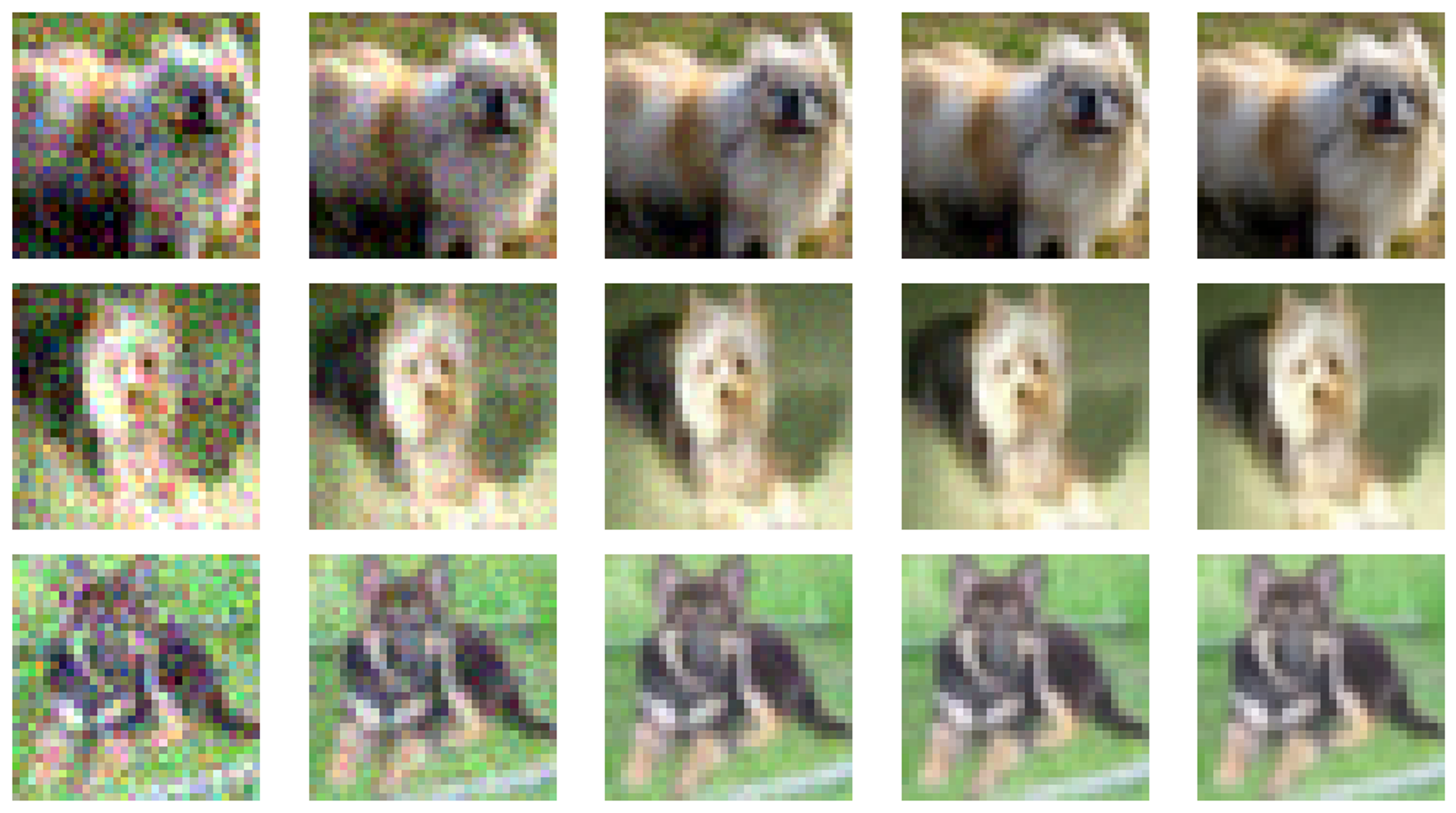}
    \captionof{figure}{Qualitative images retrieved from a pre-trained QVF under different numbers of circuit shots. From left to right, shot counts are $100$, $500$, $10^3$ and $10^4$, respectively. The rightmost images represent the ground truth.} 
    \vspace{-15pt}
    \label{samples}
\end{wrapfigure} QINR of images under finite sampling is fundamentally governed by the statistical uncertainty inherent to quantum measurement. 
The image quality depends on the number of shots, i.e.,~QINR query repetitions. 
We visualise the influence of the number of shots $N_{\text{shot}}$ (in total per image) in our ansatz in Fig.~\ref{samples} through the progressive reduction of shot noise artefacts for an increasing 
number of shots from $100$ to $10^4$. 
With low measurement shots, zero-mean sampling noise dominates the representation. 
As the number of shots $N_{\text{shot}}$ increases and approaches $10^3$, noise suppression becomes significant, as expected according to the CLT, as noise variance follows $\sigma^2 \propto 1/N_{\text{shots}}$, allowing the representation to better approximate the ground truth.
The observed noise patterns across the different shot numbers are characteristic of the proposed ansatz of QVF and will serve as a reference for future research. 

\section{Application: Image Inpainting} \label{image_inpainting} 

\setcounter{figure}{12}

\begin{wrapfigure}{r}{0.5\textwidth}
    \vspace{-10pt}
    \centering
    \includegraphics[width = 0.5\textwidth]{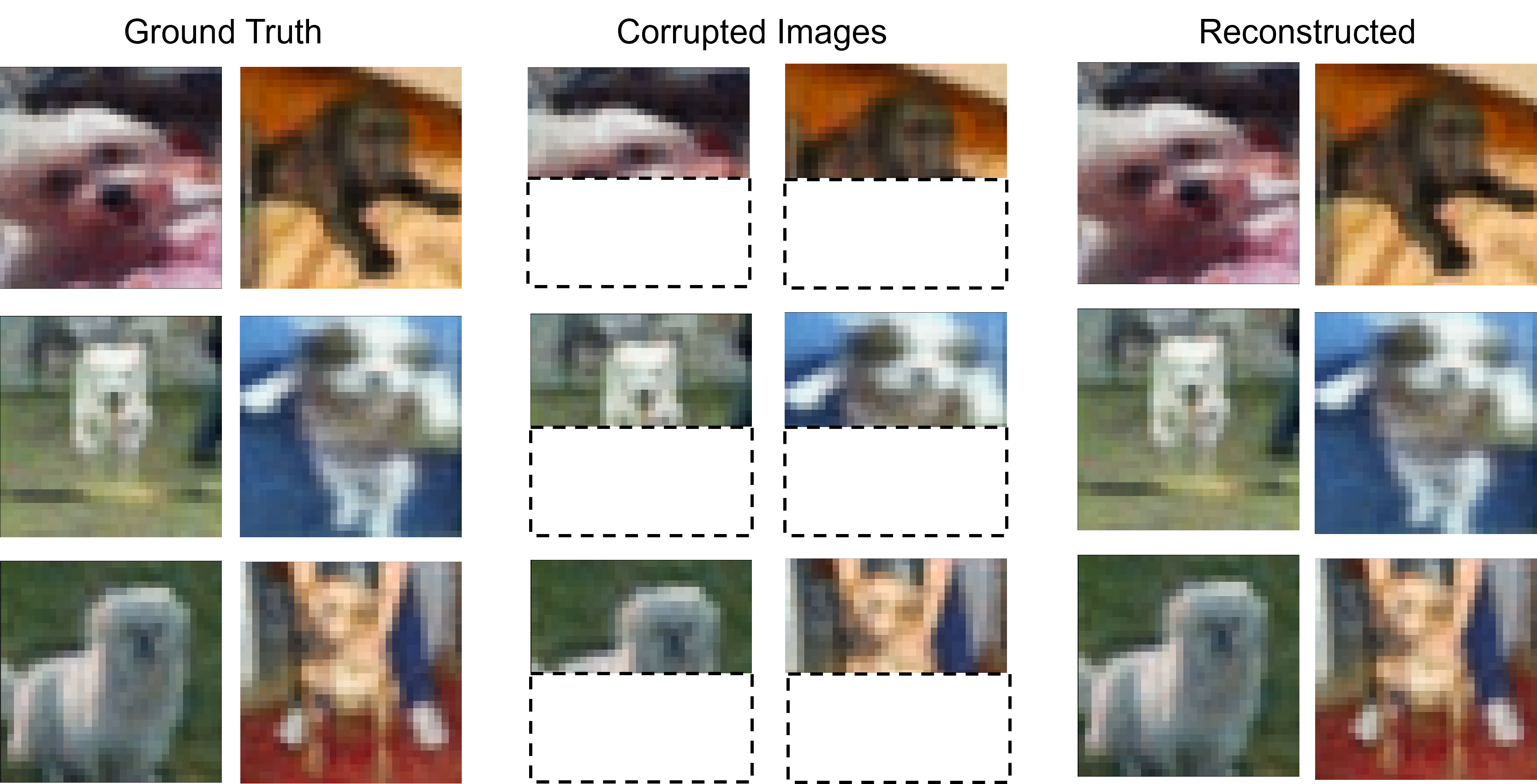} 
    \caption{Image inpainting results with QVF.}
    \label{Images}
    \vspace{-5pt}
\end{wrapfigure} 

QIREN~\cite{zhao2024quantum} and 3D-QAE~\cite{rathi20233d} are constrained by their reliance on fixed latent representations or rigid interpolation mechanisms, thereby being incapable of reconstructing complete, coherent outputs from partial or corrupted inputs.
QVF addresses such limitations by conditioning the quantum circuit topology on both the query point and a dynamic latent space, enabling applications such as image inpainting.
Given images with occluded or corrupted pixels, the circuit identifies a vector in the latent space that minimises the discrepancy between the predicted multi-dimensional properties learned by the quantum circuit and the observed noisy values. 
The optimised latent vector conditioning the quantum circuit enables recovery of missing field properties. 
Empirically, we masked out half of the image pixels and reconstructed the complete images via the protocol.
Representative results of image inpainting with QVF pre-trained on $50$ images are visualised in Fig.~\ref{Images}, demonstrating that QVF can deliver promising performance and accurately recover images even under such extreme sparsity, positioning QVF as a promising quantum circuit architecture for these tasks. 

\section{Application: Shape Completion from Partial and Noisy Depth Maps } \label{s:Noisy_shape_completion}

\noindent Similarly to image inpainting, QVF can be used for tasks such as 3D geometry completion given noisy depth maps. 
We adopt a similar setting as for images by cropping half of the samples along the depth dimension. 
We then study the effects of zero-mean Gaussian noise applied to the depth maps across different perturbation ratios $\alpha$; see Fig.~\ref{chairs}-(b). 
Shape completion performance is quantified across incremental perturbation ratios, parametrised as $\alpha \in \{0, 0.005, 0.01, 0.02, 0.03\}$, where $\alpha=0$ corresponds to an idealised noise-free scenario. 
This qualitative analysis reveals a monotonic decline in reconstruction fidelity with increasing noise, as evidenced by progressive geometric distortions and surface irregularities in Fig.~\ref{Noisy_shape_completion}. 

\section{Additional 3D Geometry Visualisations}  \label{additional_3D} 

\setcounter{figure}{11}

\begin{figure*}[t]
    \vspace{-10pt}
    \centering
    \includegraphics[width = 1.0\textwidth]{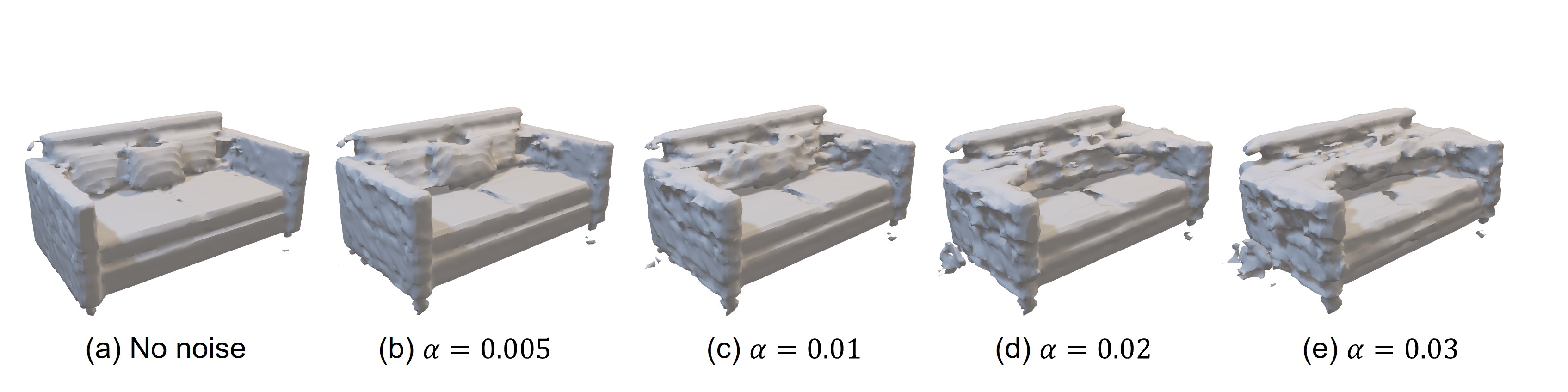}
    \caption{Shape completion from partial and noisy input depth maps using QVF; $\alpha$ is the noise ratio.}
    \label{Noisy_shape_completion}
    \vspace{-10pt}
\end{figure*}

\begin{figure*}[ht]
    \centering
    \includegraphics[width = 1.0\textwidth]{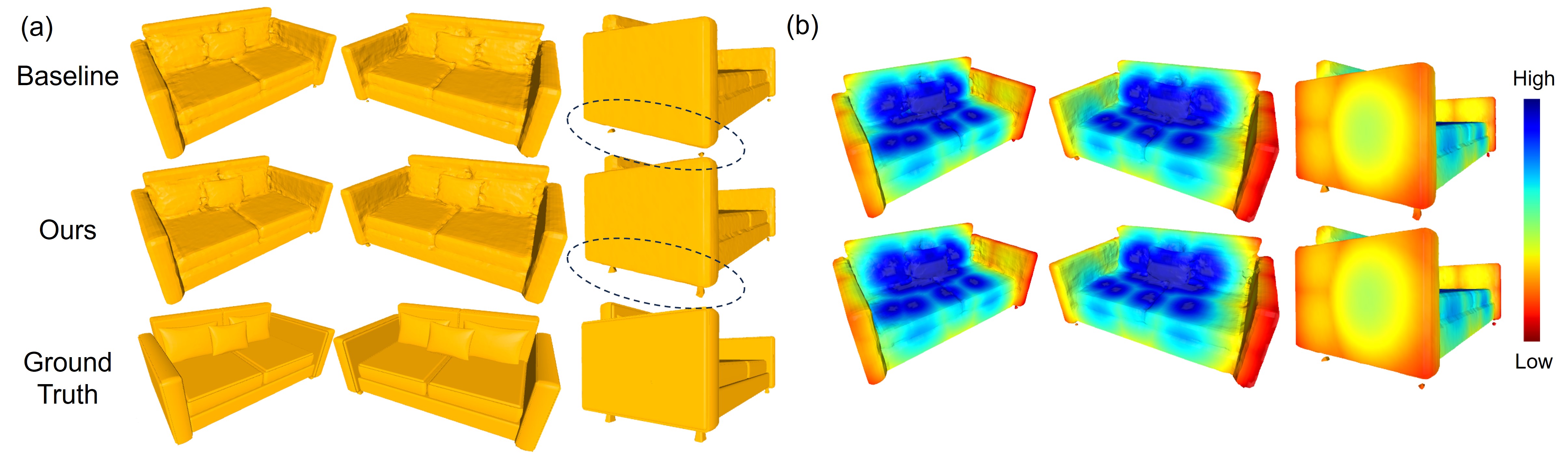} 
    \caption{(a) Comparison of geometric representations using QVF and the classical model; ground truth is presented at the bottom; (b) representation fidelity visualised via colour-encoded Hausdorff distance map: colours represent the distance to the ground truth. 
    The rendered image employs a colour gradient (blue$>$green$>$yellow$>$red) to indicate descending Hausdorff distance levels.}
    \label{more_sofas_1}
\end{figure*}

\noindent In conjunction with the quantitative results summarised in Tab.~\ref{comp_classical}, Fig.~\ref{more_sofas_1}-(a) provides a qualitative comparison of the reconstructed 3D geometries, contrasting the baseline—the classical architectural component in our model—with QVF.
The baseline’s numerically elevated loss values correlate with visual structural discontinuities, exemplified by the fragmented sofa leg, underscoring its propensity for topological inconsistencies during reconstruction.
QVF, instead, demonstrates enhanced structural coherence, generating topologically intact geometries devoid of visible artefacts, as evidenced by its preservation of fine-grained features.
Further geometric analysis, illustrated via colour-encoded per-surface Hausdorff distance distributions in Fig.~\ref{more_sofas_1}-(b), reveals systematic geometric deviations for the baseline (top) and ours (bottom), corroborating its geometric fidelity.
An interesting observation is that the colour distributions between models align well, meaning that QVF inherits the representative expressiveness of the classical component but enhances it due to inherent spectral connections to the quantum circuit. 

\section{Existing Gate-based Quantum Hardware} 
\label{ss:gate-based hardware}

Same as the prior work \cite{zhao2024quantum}, we evaluate the proposed QVF on a simulator \cite{bergholm2018pennylane} due to the immaturity of real quantum hardware for high-level and practical visual computing tasks. 
Existing gate-based quantum platforms---including superconducting circuits, trapped ions, neutral atoms, photonic systems, and quantum dots---are in varying stages of development, with none yet achieving the maturity required for large-scale, fault-tolerant computation. 
Limiting factors include noise susceptibility, restricted execution time due to quantum decoherence, and the necessity for error correction. 
Nevertheless, as classical machine learning systems continue to expand in scale, driving unprecedented computational and energy demands, rapid advancements in quantum computing techniques and hardware \cite{google_willow} are anticipated to address these barriers in the foreseeable future, underscoring the need to proactively explore applications executable on emerging quantum computers such as QVF.

\end{document}